# Bias-Driven Revision of Logical Domain Theories

**Moshe Koppel**                                                KOPPEL@BIMACS.CS.BIU.AC.IL
**Ronen Feldman**                                               FELDMAN@BIMACS.CS.BIU.AC.IL
*Department of Mathematics and Computer Science, Bar-Ilan University,*
*Ramat-Gan, Israel*

**Alberto Maria Segre**                                         SEGRE@CS.CORNELL.EDU
*Department of Computer Science, Cornell University,*
*Ithaca, NY 14853, USA*

## Abstract

The theory revision problem is the problem of how best to go about revising a deficient domain theory using information contained in examples that expose inaccuracies. In this paper we present our approach to the theory revision problem for propositional domain theories. The approach described here, called PTR, uses probabilities associated with domain theory elements to numerically track the "flow" of proof through the theory. This allows us to measure the precise role of a clause or literal in allowing or preventing a (desired or undesired) derivation for a given example. This information is used to efficiently locate and repair flawed elements of the theory. PTR is proved to converge to a theory which correctly classifies all examples, and shown experimentally to be fast and accurate even for deep theories.

## 1. Introduction

One of the main problems in building expert systems is that models elicited from experts tend to be only approximately correct. Although such hand-coded models might make a good first approximation to the real world, they typically contain inaccuracies that are exposed when a fact is asserted that does not agree with empirical observation. The *theory revision problem* is the problem of how best to go about revising a knowledge base on the basis of a collection of examples, some of which expose inaccuracies in the original knowledge base. Of course, there may be many possible revisions that sufficiently account for all of the observed examples; ideally, one would find a revised knowledge base which is both consistent with the examples and as faithful as possible to the original knowledge base.

Consider, for example, the following simple propositional domain theory, T. This theory, although flawed and incomplete, is meant to recognize situations where an investor should buy stock in a soft drink company.

*buy-stock ← increased-demand ∧ ¬product-liability*
*product-liability ← popular-product ∧ unsafe-packaging*
*increased-demand ← popular-product ∧ established-market*
*increased-demand ← new-market ∧ superior-flavor.*

The theory T essentially states that buying stock in this company is a good idea if demand for its product is expected to increase and the company is not expected to face product liability lawsuits. In this theory, product liability lawsuits may result if the product is popular (and therefore may present an attractive target for sabotage) and if the packaging is not tamper-proof. Increased product demand results if the product is popular and enjoys a large market share, or if there are new market opportunities and the product boasts a superior flavor. Using the closed world assumption, *buy-stock* is derivable given that the set of true observable propositions is precisely, say,

{*popular-product, established-market, celebrity-endorsement*}, or
{*popular-product, established-market, colorful-label*}

but not if they are, say,

{*unsafe-packaging, new-market*}, or
{*popular-product, unsafe-packaging, established-market*}.

Suppose now that we are told for various examples whether *buy-stock* should be derivable. For example, suppose we are told that if the set of true observable propositions is:

(1)     {*popular-product, unsafe-packaging, established-market*} then *buy-stock* is false,

(2)     {*unsafe-packaging, new-market*} then *buy-stock* is true,

(3)     {*popular-product, established-market, celebrity-endorsement*} then *buy-stock* is true,

(4)     {*popular-product, established-market, superior-flavor*} then *buy-stock* is false,

(5)     {*popular-product, established-market, ecologically-correct*} then *buy-stock* is false, and

(6)     {*new-market, celebrity-endorsement*} then *buy-stock* is true.

Observe that examples 2, 4, 5 and 6 are misclassified by the current theory T. Assuming that the explicitly given information regarding the examples is correct, the question is how to revise the theory so that all of the examples will be correctly classified.

## 1.1. Two Paradigms

One approach to this problem consists of enumerating partial proofs of the various examples in order to find a minimal set of domain theory elements (i.e., literals or clauses) the repair of which will satisfy all the examples (EITHER, Ourston & Mooney, in press). One problem with this approach is that proof enumeration even for a *single* example is potentially exponential in the size of the theory. Another problem with this approach is that it is unable to handle negated internal literals, and is restricted to situations where each example must belong to one and only one class. These problems suggest that it would be worthwhile to circumvent proof enumeration by employing incremental numerical schemes for focusing blame on specific elements.

A completely different approach to the revision problem is based on the use of neural networks (KBANN, Towell & Shavlik, 1993). The idea is to transform the original domain theory into network form, assigning weights in the graph according to some pre-established scheme. The connection weights are then adjusted in accordance with the observed examples using standard neural-network backpropagation techniques. The resulting network is then translated back into clausal form. The main disadvantage of this method that it lacks *representational transparency*; the neural network representation does not preserve the structure of the original knowledge base while revising it. As a result, a great deal of structural information may be lost translating back and forth between representations. Moreover, such translation imposes the limitations of both representations; for example, since neural networks are typically slow to converge, the method is practical for only very shallow domain theories. Finally, revised domain theories obtained via translation from neural networks tend to be significantly larger than their corresponding original domain theories.

Other approaches to theory revision which are much less closely related to the approach we will espouse here are RTLS (Ginsberg, 1990), KR-FOCL (Pazzani & Brunk, 1991), and ODYSSEUS (Wilkins, 1988).

## 1.2. Probabilistic Theory Revision

Probabilistic Theory Revision (PTR) is a new approach to theory revision which combines the best features of the two approaches discussed above. The starting point for PTR is the observation that any method for choosing among several possible revisions is based on some implicit bias, namely the a priori probability that each element (clause or literal) of the domain theory requires revision.

In PTR this bias is made explicit right from the start. That is, each element in the theory is assigned some a priori probability that it is not flawed. These probabilities might be assigned by an expert or simply chosen by default.

The mere existence of such probabilities solves two central problems at once. First, these probabilities very naturally define the "best" (i.e., most probable) revision out of a given set of possible revisions. Thus, our objective is well-defined; there is no need to impose artificial syntactic or semantic criteria for identifying the optimal revision. Second, these probabilities can be adjusted in response to newly-obtained information. Thus they provide a framework for incremental revision of the flawed domain theory.

Briefly, then, PTR is an algorithm which uses a set of provided examples to incrementally adjust probabilities associated with the elements of a possibly-flawed domain theory in order to find the "most probable" set of revisions to the theory which will bring it into accord with the examples.[1] Like KBANN, PTR incrementally adjusts weights associated with domain theory elements; like EITHER, all stages of PTR are carried out within the symbolic logic framework and the obtained theories are not probabilistic.

As a result PTR has the following features:

(1)   it can handle a broad range of theories including those with negated internal literals and multiple roots.

(2)   it is linear in the size of the theory times the number of given examples.

(3)   it produces relatively small, accurate theories that retain much of the structure of the original theory.

(4)   it can exploit additional user-provided bias.

In the next section of this paper we formally define the theory revision problem and discuss issues of data representation. We lay the foundations for any future approach to theory revision by introducing very sharply defined terminology and notation. In Section 3 we propose an efficient algorithm for finding flawed elements of a theory, and in Section 4 we show how to revise these elements. Section 5 describes how these two components are combined to form the PTR algorithm. In Section 5, we also discuss the termination and convergence properties of PTR and walk through a simple example of PTR in action. In Section 6 we experimentally evaluate PTR and compare it to other theory revision algorithms. In Section 7, we sum up our results and indicate directions for further research.

_______________________

[1] In the following section we will make precise the notion of "most probable set of revisions."

The formal presentation of the work described here is, unfortunately, necessarily dense. To aid the more casual reader, we have moved all formal proofs to three separate appendices. In particular, in the third appendix we prove that, under appropriate conditions, PTR converges. Reading of these appendices can safely be postponed until after the rest of the paper has been read. In addition, we provide in Appendix D, a "quick reference guide" to the notation used throughout the paper. We would suggest that a more casual reader might prefer to focus on Section 2, followed by a cursory reading of Sections 3 and 4, and a more thorough reading of Section 5.

## 2. Representing the Problem

A *propositional domain theory*, denoted $\Gamma$, is a stratified set of clauses of the form $C_i\colon H_i \leftarrow B_i$ where $C_i$ is a clause label, $H_i$ is a proposition (called the *head* of $C_i$) and $B_i$ is a set of positive and negative literals (called the *body* of $C_i$). As usual, the clause $C_i\colon H_i \leftarrow B_i$ represents the assertion that the proposition $H_i$ is implied by the conjunction of literals in $B_i$. The domain theory is simply the conjunction of its clauses. It may be convenient to think of this as a propositional logic program without facts (but with negation allowed).

A proposition which does not appear in the head of any clause is said to be *observable*. A proposition which appears in the head of some clause but does not appear in the body of any clause is called a *root*. An *example, E*, is a truth assignment to all observable propositions. It is convenient to think of $E$ as a set of true observable propositions.

Let $\Gamma$ be a domain theory with roots $r_1, \cdots, r_n$. For an example, $E$, we define the vector $\Gamma(E) = \langle \Gamma_1(E), \cdots, \Gamma_n(E) \rangle$ where $\Gamma_i(E) = 1$ if $E \vdash_\Gamma r_i$ (using resolution) and $\Gamma_i(E) = 0$ if $E \not\vdash_\Gamma r_i$. Intuitively, $\Gamma(E)$ tells us which of the conclusions $r_1, \cdots, r_n$ can be drawn by the expert system when given the truth assignment $E$.

Let the *target* domain theory, $\Theta$, be some domain theory which accurately models the domain of interest. In other words, $\Theta$ represents the correct domain theory. An ordered pair, $\langle E, \Theta(E) \rangle$, is called an *exemplar* of the domain: if $\Theta_i(E) = 1$ then the exemplar is said to be an *IN exemplar* of $r_i$, while if $\Theta_i(E) = 0$ then the exemplar is said to be an *OUT exemplar* of $r_i$. Typically, in theory revision, we know $\Theta(E)$ without knowing $\Theta$.

Let $\Gamma$ be some possibly incorrect theory for a domain which is in turn correctly modeled by the target theory $\Theta$. Any inaccuracies in $\Gamma$ will be reflected by exemplars for which $\Gamma(E) \neq \Theta(E)$. Such exemplars are said to be *misclassified* by $\Gamma$. Thus, a *misclassified IN exemplar for $r_i$*, or *false negative for $r_i$*, will have $\Theta_i(E) = 1$ but $\Gamma_i(E) = 0$, while a *misclassified OUT exemplar for $r_i$*, or *false positive for $r_i$*, will have $\Theta_i(E) = 0$ but $\Gamma_i(E) = 1$.[2] Typically, in theory revision we know $\Theta(E)$ without knowing $\Theta$.

Consider, for example, the domain theory, $T$, and example set introduced in Section 1. The theory $T$ has only a single root, *buy-stock*. The observable propositions mentioned in the examples are *popular-product*, *unsafe-packaging*, *established-market*, *new-market*, *celebrity-*

---

[2] We prefer the new terminology "IN/OUT" to the more standard "positive/negative" because the latter is often used to refer to the classification of the example by the given theory, while we use "IN/OUT" to refer specifically to the actual classification of the example.

*endorsement*, *superior-flavor*, and *ecologically-correct*. For the example $E = \{ \textit{unsafe-packaging}, \textit{new-market} \}$ we have $\mathrm{T}(E) = \langle \mathrm{T}_1(E) \rangle = \langle 0 \rangle$. Nevertheless, we are told that $\Theta(E) = \langle \Theta_1(E) \rangle = \langle 1 \rangle$. Thus, $E = \langle \{ \textit{unsafe-packaging}, \textit{new-market} \}, \langle 1 \rangle \rangle$ is a misclassified IN exemplar of the root *buy-stock*.

Now, given misclassified exemplars, there are four *revision operators* available for use with propositional domain theories:

(1)    add a literal to an existing clause,

(2)    delete an existing clause,

(3)    add a new clause, and

(4)    delete a literal from an existing clause.

For negation-free domain theories, the first two operations result in *specializing* $\Gamma$, since they may allow some IN exemplars to become OUT exemplars. The latter two operations result in *generalizing* $\Gamma$, since they may allow some OUT exemplars to become IN exemplars.[3]

We say that a set of revisions to $\Gamma$ is *adequate* for a set of exemplars if, after the revision operators are applied, all the exemplars are correctly classified by the revised domain theory $\Gamma'$. Note that we are not implying that $\Gamma'$ is identical to $\Theta$, but rather that for every exemplar $\langle E, \Theta(E) \rangle$, $\Gamma'(E) = \Theta(E)$. Thus, there may be more than one adequate revision set. The goal of any theory revision system, then, is to find the "best" revision set for $\Gamma$, which is adequate for a given set of exemplars.

## 2.1. Domain Theories as Graphs

In order to define the problem even more precisely and to set the stage for its solution, we will show how to represent a domain theory in the form of a weighted digraph. We begin by defining a more general version of the standard AND–OR proof tree, which collapses the distinction between AND nodes and OR nodes.

For any set of propositions $\{ P_1, \cdots, P_n \}$, let $\mathrm{NAND}(\{ P_1, \cdots, P_n \})$ be a Boolean formula which is false if and only if $\{ P_1, \cdots, P_n \}$ are all true. Any domain theory $\Gamma$ can be translated into an equivalent domain theory $\hat{\Gamma}$ consisting of NAND equations as follows:

(1)    For each clause $C_i \colon H_i \leftarrow B_i \in \Gamma$, the equation $\hat{C}_i = \mathrm{NAND}(B_i)$ is in $\hat{\Gamma}$.

(2)    For each non-observable proposition $P$ appearing in $\Gamma$ the equation $P = \mathrm{NAND}(C_P)$ is in $\hat{\Gamma}$, where $C_P = \{ \hat{C}_i \mid H_i = P \}$, i.e., the set consisting of the label of each clause in $\Gamma$ whose head is $P$.

(3)    For each negative literal $\neg P$ appearing in $\Gamma$, the equation $\neg P = \mathrm{NAND}(\{ P \})$ is in $\hat{\Gamma}$.

$\hat{\Gamma}$ contains no equations other than these. Observe that the literals of $\hat{\Gamma}$ are the literals of $\Gamma$ together with the new literals $\{ \hat{C}_i \}$ which correspond to the clauses of $\Gamma$. Most important, $\hat{\Gamma}$ is equivalent to $\Gamma$ in the sense that for each literal $l$ in $\Gamma$ and any assignment $E$ of truth values to the observable propositions of $\Gamma$, $E \mid\!\!- _{\hat{\Gamma}} l$ if and only if $E \mid\!\!- _\Gamma l$.

---

[3] In the event that negative literals appear in the domain theory, the consequences of applying these operators are slightly less obvious. This will be made precise in the second part of this section.

Consider, for example, the domain theory T of Section 1. The set of NAND equations $\hat{\mathrm{T}}$ is

$buy\text{-}stock = \mathrm{NAND}(\{C_1\})$,
$C_1 = \mathrm{NAND}(\{increased\text{-}demand,\ \neg product\text{-}liability\})$,
$\neg product\text{-}liability = \mathrm{NAND}(\{product\text{-}liability\})$,
$increased\text{-}demand = \mathrm{NAND}(\{C_3, C_4\})$,
$product\text{-}liability = \mathrm{NAND}(\{C_2\})$,
$C_2 = \mathrm{NAND}(\{popular\text{-}product,\ unsafe\text{-}packaging\})$,
$C_3 = \mathrm{NAND}(\{popular\text{-}product,\ established\text{-}market\})$, and
$C_4 = \mathrm{NAND}(\{new\text{-}market,\ superior\text{-}flavor\})$.

Observe that $buy\text{-}stock$ is true in $\hat{\mathrm{T}}$ for precisely those truth assignments to the observables for which $buy\text{-}stock$ is true in $T$.

We now use $\hat{\Gamma}$ to obtain a useful graph representation of $\Gamma$. For an equation $\hat{\Gamma}_i$ in $\hat{\Gamma}$, let $h(\hat{\Gamma}_i)$ refer to the left side of $\hat{\Gamma}_i$ and let $b(\hat{\Gamma}_i)$ refer to the set of literals which appear on the right side of $\hat{\Gamma}_i$. In other words, $h(\hat{\Gamma}_i) = \mathrm{NAND}(b(\hat{\Gamma}_i))$.

>**Definition**: A *dt-graph* $\Delta_\Gamma$ for a domain theory $\Gamma$ consists of a set of nodes which
>correspond to the literals of $\hat{\Gamma}$ and a set of directed edges corresponding to the set
>of ordered pairs $\{\langle x, y \rangle \mid x = h(\hat{\Gamma}_i), y \in b(\hat{\Gamma}_i), \hat{\Gamma}_i \in \hat{\Gamma}\}$. In addition, for each root
>$r$ we add an edge, $e_r$, leading into $r$ (from some artificial node).

In other words, $\Delta_\Gamma$ consists of edges from each literal in $\hat{\Gamma}$ to each of its antecedents. The dt-graph representation of T is shown in Figure 1.

Let $n_e$ be the node to which the edge $e$ leads and let $n^e$ be the node from which it comes. If $n_e$ is a clause, then we say that $e$ is a *clause edge*; if $n_e$ is a root, then we say that $e$ is a *root edge*; if $n_e$ is a literal and $n^e$ is a clause, then we say that $e$ is a *literal edge*; if $n_e$ is a proposition and $n^e$ is its negation, then we say that $e$ is a *negation edge*.

The dt-graph $\Delta_\Gamma$ is very much like an AND–OR graph for $\Gamma$. It has, however, a very significant advantage over AND–OR graphs because it collapses the distinction between clause edges and literal edges which is central to the AND–OR graph representation. In fact, even negation edges (which do not appear at all in the AND–OR representation) are not distinguished from literal edges and clause edges in the dt-graph representation.

In terms of the dt-graph $\Delta_\Gamma$, there are two basic revision operators — deleting edges or adding edges. What are the effects of adding or deleting edges from $\Delta_\Gamma$? If the length of every path from a root $r$ to a node $n$ is even (odd) then $n$ is said to be an even (odd) node for $r_i$. If $n^e$ is even (odd) for $r_i$, then $e$ is said to be even (odd) for $r_i$. (Of course it is possible that the depth of an edge is neither even nor odd.) Deleting an even edge for $r_i$ *specializes* the definitions of $r_i$ in the sense that if $\Delta_{\Gamma'}$ is the result of the deletion, then $\Gamma'_i(E) \leq \Gamma_i(E)$ for all exemplars $\langle E, \Theta(E) \rangle$; likewise, adding an even edge for $r_i$ generalizes the definition of $r_i$ in the sense that if $\Delta_{\Gamma'}$ is the result of adding the edge to $\Delta_\Gamma$ then $\Gamma'_i(E) \geq \Gamma_i(E)$. Analogously, deleting an odd edge for $r_i$ generalizes the definition of $r_i$, while adding an odd edge for $r_i$ specializes the definition of $r_i$. (Deleting or adding an edge which is neither odd nor even for $r_i$ might result in a new definition of $r_i$ which is neither strictly more general nor strictly more specific.)

To understand this intuitively, first consider the case in which there are no negation edges in $\Delta_\Gamma$. Then an even edge in $\Delta_\Gamma$ represents a clause in $\Gamma$, so that deleting is specialization and adding is generalization. An odd edge in $\Delta_\Gamma$ represents a literal in the body of a clause in $\Gamma$ so that deleting is generalization and adding a specialization. Now, if an odd number of negation edges

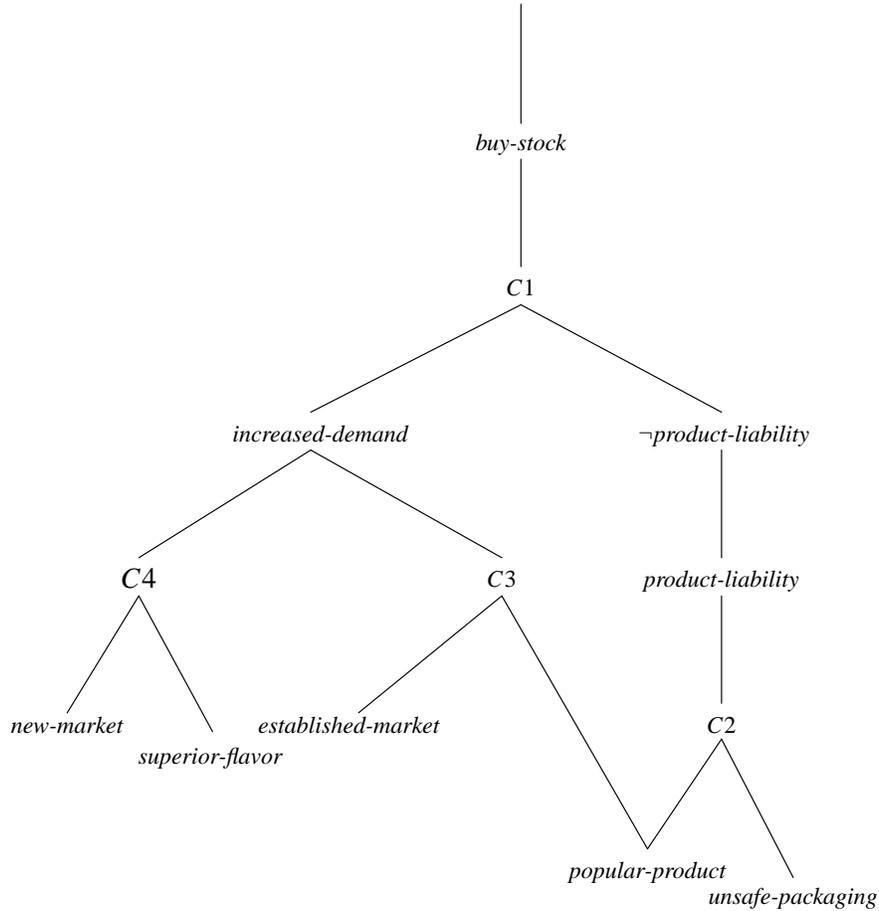

Figure 1: The dt-graph, $\Delta_T$, of the theory T.

are present on the path from $r_i$ to an edge then the role of the edge is reversed.

## 2.2. Weighted Graphs

A *weighted dt-graph* is an ordered pair $\langle \Delta_\Gamma, w \rangle$ where $\Delta_\Gamma$ is a dt-graph $w$ and is an assignment of values in $(0, 1]$ to each node and edge in $\Delta_\Gamma$. For an edge $e$, $w(e)$ is meant to represent the user's degree of confidence that the edge $e$ need not be deleted to obtain the correct domain theory. For a node $n$, $w(n)$ is the user's degree of confidence that no edge leading from the node $n$ need be added in order to obtain the correct domain theory. Thus, for example, the assignment $w(n) = 1$ means that it is certain that no edge need be added to the node $n$ and the assignment $w(e)$ means that it is certain that $e$ should not be deleted. Observe that if the node $n$ is labeled by a negative literal or an observable proposition then $w(n) = 1$ by definition, since graphs obtained by adding edges to such nodes do not correspond to any domain theory. Likewise, if $e$ is a root-edge or a negation-edge, then $w(e) = 1$.

For practical reasons, we conflate the weight $w(e)$ of an edge $e$ and the weight, $w(n_e)$, of the node $n_e$, into a single value, $p(e) = w(e) \times w(n_e)$, associated with the edge $e$. The value $p(e)$ is the user's confidence that $e$ need not be repaired, either by deletion or by dilution via addition of child edges.

There are many ways that these values can be assigned. Ideally, they can be provided by the expert such that they actually reflect the expert's degree of confidence in each element of the theory. However, even in the absence of such information, values can be assigned by default; for example, all elements can be assigned equal value. A more sophisticated method of assigning values is to assign higher values to elements which have greater "semantic impact" (e.g., those closer to the roots). The details of one such method are given in Appendix A. It is also, of course, possible for the expert to assign some weights and for the rest to be assigned according to some default scheme. For example, in the weighted dt-graph, $\langle \Delta_T, p \rangle$, shown in Figure 2, some edges have been assigned weight near 1 and others have been assigned weights according to a simple default scheme.

The semantics of the values associated with the edges can be made clear by considering the case in which it is known that the correct dt-graph is a subset of the given dt-graph, $\Delta$. Consider a probability function on the space of all subgraphs of $\Delta$. The weight of an edge is simply the sum of the probabilities of the subgraphs in which the edge appears. Thus the weight of an edge is the probability that the edge does indeed appear in the target dt-graph. We easily extend this to the case where the target dt-graph is not necessarily a subgraph of the given one.[4]

Conversely, given only the probabilities associated with edges and assuming that the deletion of different edges are independent events, we can compute the probability of a subgraph, $\Delta'$. Since $p(e)$ is the probability that $e$ is not deleted and $1 - p(e)$ is the probability that $e$ is deleted, it follows that

$$p(\Delta') = \prod_{e \in \Delta'} p(e) \times \prod_{e \in \Delta - \Delta'} 1 - p(e).$$

Letting $S = \Delta - \Delta'$, we rewrite this as

$$p(\Delta') = \prod_{e \in \Delta - S} p(e) \times \prod_{e \in S} 1 - p(e).$$

We use this formula as a basis for assigning a value to each dt-graph $\Delta'$ obtainable from $\Delta$ via revision of the set of edges $S$, even in the case where edge-independence does not hold and even in the case in which $\Delta'$ is not a subset of $\Delta$. We simply define

$$w(\Delta') = \prod_{e \in \Delta - S} p(e) \times \prod_{e \in S} 1 - p(e).$$

(In the event that $\Delta$ and $\Delta'$ are such that $S$ is not uniquely defined, choose $S$ such that $w(\Delta')$ is maximized.) Note that where independence holds and $\Delta'$ is subgraph of $\Delta$, we have

---

[4] In order to avoid confusion it should be emphasized that the meaning of the weights associated with edges is completely different than that associated with edges of Pearl's Bayesian networks (1988). For us, these weights represent a meta-domain-theory concept: the probability that this edge appears in some unknown target domain theory. For Pearl they represent conditional probabilities within a probabilistic domain-theory. Thus, the updating method we are about to introduce is totally unrelated to that of Pearl.

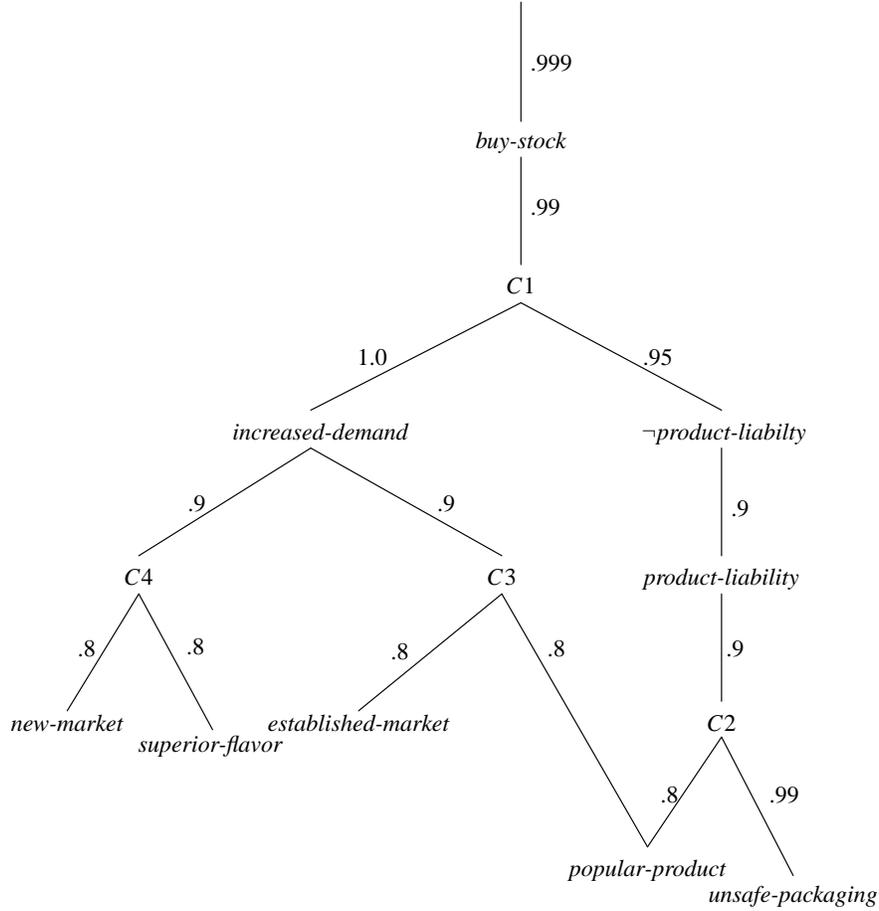

Figure 2: The weighted dt-graph, $\langle \Delta_T, p \rangle$.

$w(\Delta') = p(\Delta')$.

## 2.3. Objectives of Theory Revision

Now we can formally define the proper objective of a theory revision algorithm:

> *Given a weighted dt-graph $\langle \Delta, p \rangle$ and a set of exemplars Z, find a dt-graph $\Delta'$ such that $\Delta'$ correctly classifies every exemplar in Z and $w(\Delta')$ is maximal over all such dt-graphs.*

Restating this in the terminology of information theory, we define the *radicality* of a dt-graph $\Delta'$ relative to an initial weighted dt-graph $K = \langle \Delta, p \rangle$ as

$$Rad_K(\Delta') = \sum_{e \in \Delta-S} -log(p(e)) + \sum_{e \in S} -log(1-p(e))$$

where $S$ is the set of edges of $\Delta$ which need to be revised in order to obtain $\Delta'$. Thus given a weighted dt-graph K and a set of exemplars Z, we wish to find the least radical dt-graph relative to K which correctly classifies the set of exemplars Z.

Note that radicality is a straightforward measure of the quality of a revision set which neatly balances syntactic and semantic considerations. It has been often noted that minimizing syntactic change alone can lead to counter-intuitive results by giving preference to changes near the root which radically alter the semantics of the theory. On the other hand, regardless of the distribution of examples, minimizing semantic change alone results in simply appending to the domain theory the correct classifications of the given misclassified examples without affecting the classification of any other examples.

Minimizing radicality automatically takes into account both these criteria. Thus, for example, by assigning higher initial weights to edges with greater semantic impact (as in our default scheme of Appendix A), the syntactic advantage of revising close to the root is offset by the higher cost of such revisions. For example, suppose we are given the theory T of the introduction and the single misclassified exemplar

$$\langle \{ \textit{unsafe-packaging}, \textit{new-market} \}, \langle 1 \rangle \rangle .$$

There are several possible revisions which would bring T into accord with the exemplar. We could, for example, add a new clause

$$\textit{buy-stock} \leftarrow \textit{unsafe-packaging} \wedge \textit{new-market},$$

delete *superior-flavor* from clause $C4$, delete *popular-product* and *established-market* from clause $C3$, or delete *increased-demand* from clause $C1$. Given the weights of Figure 2, the deletion of *superior-flavor* from clause $C4$ is clearly the least radical revision.

Observe that in the special case where all edges are assigned identical initial weights, regardless of their semantic strength, minimization of radicality does indeed reduce to a form of minimization of syntactic change. We wish to point out, however, that even in this case our definition of "syntactic change" differs from some previous definitions (Wogulis & Pazzani, 1993). Whereas those definitions count the number of deleted and added edges, we count the number of edges deleted or added *to*. To understand why this is preferable, consider the case in which some internal literal, which happens to have a large definition, is omitted from one of the clause in the target theory. Methods which count the number of added edges will be strongly biased against restoring this literal, preferring instead to make *several* different repairs which collectively involve fewer edges than to make a *single* repair involving more edges. Nevertheless, given the assumption that the probabilities of the various edges in the given theory being mistaken are equal, it is far more intuitive to repair only at a single edge, as PTR does. (We agree, though, that once an edge has been chosen for repair, the chosen repair should be minimal over all equally effective repairs.)

## 3. Finding Flawed Elements

PTR is an algorithm which finds an adequate set of revisions of approximately minimum radicality. It achieves this by locating flawed edges and then repairing them. In this section we give the algorithm for locating flawed edges; in the next section we show how to repair them.

The underlying principle of locating flawed edges is to process exemplars one at a time, in each case updating the weights associated with edges in accordance with the information contained in the exemplars. We measure the "flow" of a proof (or refutation) through the edges of the graph. The more an edge contributes to the correct classification of an example, the more its weight is raised; the more it contributes to the misclassification of the example, the more its weight is lowered. If the weight of an edge drops below a prespecified revision threshold $\sigma$, it is revised.

The core of the algorithm is the method of updating the weights. Recall that the weight represents the probability that an edge appears in the target domain theory. The most natural way to update these weights, then, is to replace the probability that an edge need not be revised with the conditional probability that it need not be revised *given the classification of an exemplar*. As we shall see later, the computation of conditional probabilities ensures many desirable properties of updating which ad hoc methods are liable to miss.

### 3.1. Processing a Single Exemplar

One of the most important results of this paper is that *under certain conditions the conditional probabilities of all the edges in the graph can be computed in a single bottom-up-then-top-down sweep through the dt-graph*. We shall employ this method of computation even when those conditions do not hold. In this way, updating is performed in highly efficient fashion while, at the same time, retaining the relevant desirable properties of conditional probabilities.

More precisely, the algorithm proceeds as follows. We think of the nodes of $\Delta_\Gamma$ which represent observable propositions as input nodes, and we think of the values assigned by an example $E$ to each observable proposition as inputs. Recall that the assignment of weights to the edges is associated with an implicit assignment of probabilities to various dt-graphs obtainable via revision of $\Delta_\Gamma$. For some of these dt-graphs, the root $r_i$ is provable from the example $E$, while for others it is not. We wish to make a bottom-up pass through $K = \langle \Delta_\Gamma, p \rangle$ in order to compute (or at least approximate) for each root $r_i$, the probability that the target domain theory is such that $r_i$ is true for the example $E$. The obtained probability can then be compared with the desired result, $\Theta_i(E)$, and the resulting difference can be used as a basis for adjusting the weights, $w(e)$, for each edge $e$.

Let

$$E(P) = \begin{cases} 1 & \text{if } P \text{ is true in } E \\ 0 & \text{if } P \text{ is false in } E. \end{cases}$$

We say that a node $n \in \Delta_\Gamma$ is *true* if the literal of $\hat{\Gamma}$ which labels it is true. Now, a node passes the value "true" up the graph if it is either true or deleted, i.e., if it is not both undeleted and false. Thus, for an edge $e$ such that $n_e$ is the observable proposition $P$, the value $u_E(e) = 1 - [p(e) \times (1 - E(P))]$ is the probability of the value "true" being passed up the graph from $e$.[5]

Now, recalling that a node in $\Delta_\Gamma$ represents a NAND operation, if the truth of a node in $\Delta_\Gamma$ is independent of the truth of any of its brothers, then for any edge $e$, the probability of "true" being passed up the graph is

$$u_E(e) = 1 - p(e) \prod_{s \,\in\, children(e)} u_E(s).$$

---

[5] Note that, in principle, the updating can be performed exactly the same way even if $0 < E(P) < 1$. Thus, the algorithm extends naturally to the case in which there is uncertainty regarding the truth-value of some of the observable propositions.

We call $u_E(e)$ the *flow* of $E$ through $e$.

We have defined the flow $u_E(e)$ such that, under appropriate independence conditions, for any node $n_e$, $u_E(e)$ is in fact the probability that $n_e$ is true given $\langle \Delta_\Gamma, w \rangle$ and $E$. (For a formal proof of this, see Appendix B.) In particular, for a root $r_i$, the flow $u_E(e_{r_i})$ is, even in the absence of the independence conditions, a good approximation of the probability that the target theory is such that $r_i$ is true given $\langle \Delta_\Gamma, w \rangle$ and $E$.

In the second stage of the updating algorithm, we propagate the difference between each computed value $u_E(e_{r_i})$ (which lies somewhere between 0 and 1) and its target value $\Theta_i(E)$ (which is either 0 or 1) top-down through $\Delta_\Gamma$ in a process similar to backpropagation in neural networks. As we proceed, we compute a new value $v_E(e)$ as well as an updated value for $p(e)$, for every edge $e$ in $\Delta_\Gamma$. The new value $v_E(e)$ represents an updating of $u_E(e)$ where the correct classification, $\Theta(E)$, of the example $E$ has been taken into account.

Thus, we begin by setting each value $v_E(r_i)$ to reflect the correct classification of the example. Let $\varepsilon > 0$ be some very small constant[6] and let

$$v_E(e_{r_i}) = \begin{cases} \varepsilon & \text{if } \Theta_i(E) = 0 \\ 1 - \varepsilon & \text{if } \Theta_i(E) = 1. \end{cases}$$

Now we proceed top down through $\Delta_\Gamma$, computing $v_E(e)$ for each edge in $\Delta_\Gamma$. In each case we compute $v_E(e)$ on the basis of $u_E(e)$, that is, on the basis of how much of the proof (or refutation) of $E$ flows through the edge $e$. The precise formula is

$$v_E(e) = 1 - (1 - u_E(e)) \times \frac{v_E(f(e))}{u_E(f(e))}$$

where $f(e)$ is that parent of $e$ for which $\left| 1 - \dfrac{\max[v_E(f(e)), u_E(f(e))]}{\min[v_E(f(e)), u_E(f(e))]} \right|$ is greatest. We show in Appendix B why this formula works.

Finally, we compute $p_{new}(e)$, the new values of $p(e)$, using the current value of $p(e)$ and the values of $v_E(e)$ and $u_E(e)$ just computed:

$$p_{new}(e) = 1 - (1 - p(e)) \times \frac{v_E(e)}{u_E(e)}.$$

If the deletion of different edges are independent events and $\Theta$ is known to be a subgraph of $\Gamma$, then $p_{new}(e)$ is the conditional probability that the edge $e$ appears in $\Theta$, given the exemplar $\langle E, \Theta(E) \rangle$ (see proof in Appendix B). Figure 3 gives the pseudo code for processing a single exemplar.

---

[6] Strictly speaking, for the computation of conditional probabilities, we need to use $\varepsilon = 0$. However, in order to ensure convergence of the algorithm in all cases, we choose $\varepsilon > 0$ (see Appendix C). In the experiments reported in Section 6, we use the value $\varepsilon = .01$.

**function** *BottomUp*($\langle \Delta, p \rangle$ : *weighted dt-graph*,  *E* : *exemplar*): *array of real*;
  **begin**
    $S \Leftarrow \varnothing$ ; $V \Leftarrow Leaves(\Delta)$;
    **for** $e \in Leaves(\Delta)$ **do**
      **begin**
        **if** $e \in E$ **then** $u(e) \Leftarrow 1$;
        **else** $u(e) \Leftarrow 1 - p(e)$;
        $S \Leftarrow Merge(S, Parents(e, \Delta))$;
      **end**
    **while** $S \neq \varnothing$ **do**
      **begin**
        $e \Leftarrow PopSuitableParent(S, V)$; $V \Leftarrow AddElement(e, V)$;
        $u(e) \Leftarrow 1 - (p(e) \prod_{d \in Children(e, \Delta)} u(d))$;
        $S \Leftarrow Merge(S, Parents(e, \Delta))$;
      **end**
    **return** $u$;
  **end**

**function** *TopDown*($\langle \Delta, p \rangle$ : *weighted dt-graph*,  *u* : *array of real*,
                    *E* : *exemplar*,  $\varepsilon$ : *real*): *weighted dt-graph*;
  **begin**
    $S \Leftarrow \varnothing$ ; $V \Leftarrow Roots(\Delta)$;
    **for** $r_i \in Roots(\Delta)$ **do**
      **begin**
        **if** $\Gamma_i(E) = 1$ **then** $v(r_i) \Leftarrow \varepsilon$;
        **else** $v(r_i) \Leftarrow 1 - \varepsilon$;
        $S \Leftarrow Merge(S, Children(r_i, \Delta))$;
      **end**
    **while** $S \neq \varnothing$ **do**
      **begin**
        $e \Leftarrow PopSuitableChild(S, V)$; $V \Leftarrow AddElement(e, V)$; $f \Leftarrow MostChangedParent(e, \Delta)$;
        $v(e) \Leftarrow 1 - (1 - u(e)) \times \dfrac{v(f)}{u(f)}$ ;
        $p(e) \Leftarrow 1 - (1 - p(e)) \times \dfrac{v(e)}{u(e)}$ ;
        $S \Leftarrow Merge(S, Children(e, \Delta))$;
      **end**
    **return** $\langle \Delta, p \rangle$;
  **end**

Figure 3: Pseudo code for processing a single exemplar. The functions *BottomUp* and *TopDown* sweep the dt-graph. *BottomUp* returns an array on edges representing proof flow, while *TopDown* returns an updated weighted dt-graph. We are assuming the dt-graph datastructure has been defined and initialized appropriately. Functions *Children*, *Parents*, *Roots*, and *Leaves* return sets of edges corresponding to the corresponding graph relation on the dt-graph. Function *Merge* and *AddElement* operate on sets, and functions *PopSuitableParent* and *PopSuitableChild* return an element of its first argument whose children or parents, respectively, are all already elements of its second argument while simultaneously deleting the element from the first set, thus guaranteeing the appropriate graph traversal strategy.

Consider the application of this updating algorithm to the weighted dt-graph of Figure 2. We are given the exemplar ⟨{*unsafe-packaging*, *new-market*}, ⟨1⟩⟩, i.e., the example in which *unsafe-packaging* and *new-market* are true (and all other observables are false) should yield a derivation of the root *buy-stock*. The weighted dt-graph obtained by applying the algorithm is shown in Figure 4.

This example illustrates some important general properties of the method.

(1)     *Given an* IN *exemplar, the weight of an odd edge cannot decrease and the weight of an even edge cannot increase.* (The analogous property holds for an OUT exemplar.) In the case where no negation edge appears in $\Delta_\Gamma$, this corresponds to the fact that a clause cannot help prevent a proof, and literals in the body of a clause cannot help complete a proof. Note in particular that the weights of the edges corresponding to the literals *popular-product* and *established-market* in clause $C3$ dropped by the same amount, reflecting the identical roles played by them in this example. However, the weight of the edge corresponding to the literal *superior-flavor* in clause $C4$ drops a great deal more than both those edges, reflecting the fact that the deletion of *superior-flavor* alone would allow a proof of *buy-stock*, while the deletion of either *popular-product* alone or *established-market* alone would not allow a proof of *buy-stock*.

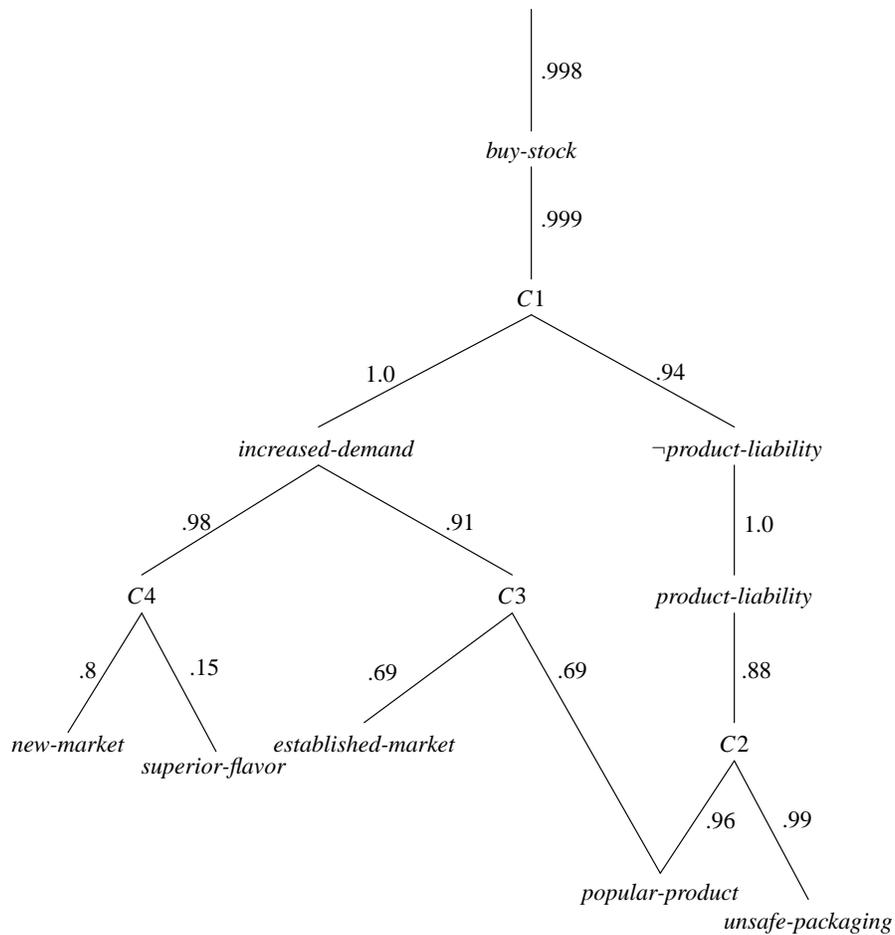

Figure 4:  The weighted dt-graph of Figure 2 after processing the exemplar ⟨{*unsafe-packaging*, *new-market*}, ⟨1⟩⟩.

(2) *An edge with initial weight 1 is immutable; its weight remains 1 forever.* Thus although an edge with weight 1, such as that corresponding to the literal *increased-demand* in clause $C1$, may contribute to the prevention of a desired proof, its weight is not diminished since we are told that there is no possibility of that literal being flawed.

(3) *If the processed exemplar can only be correctly classified if a particular edge e is revised, then the updated probability of e will approach 0 and e will be immediately revised.*[7] Thus, for example, were the initial weights of the edge corresponding to *established-market* and *popular-product* in $C3$ to approach 1, the weight of the edge corresponding to *superior-flavor* in $C4$ would approach 0. Since we use weights only as a temporary device for locating flawed elements, this property renders our updating method more appropriate for our purposes then standard backpropagation techniques which adjust weights gradually to ensure convergence.

(4) *The computational complexity of processing a single exemplar is linear in the size of the theory* $\Gamma$. Thus, the updating algorithm is quite efficient when compared to revision techniques which rely on enumerating all proofs for a root. Note further that the computation required to update a weight is identical for every edge of $\Delta_\Gamma$ regardless of edge type. Thus, PTR is well suited for mapping onto fine-grained SIMD machines.

### 3.2. Processing Multiple Exemplars

As stated above, the updating method is applied iteratively to one example at a time (in random order) until some edge drops below the revision threshold, $\sigma$. If after a complete cycle no edge has dropped below the revision threshold, the examples are reordered (randomly) and the updating is continued.[8]

For example, consider the weighted dt-graph of Figure 2. After processing the exemplars

⟨ {*unsafe-packaging*, *new-market*}, ⟨ 1 ⟩ ⟩,
⟨ {*popular-product*, *established-market*, *superior-flavor*}, ⟨ 0 ⟩ ⟩, and
⟨ {*popular-product*, *established-market*, *celebrity-endorsement*}, ⟨ 0 ⟩ ⟩

we obtain the dt-graph shown in Figure 5. If our threshold is, say, $\sigma = .1$, then we have to revise the edge corresponding to the clause $C3$. This reflects the fact that the clause $C3$ has contributed substantially to the misclassification of the second and third examples from the list above while not contributing substantially to the correct classification of the first.

---

[7] If we were to choose $\varepsilon = 0$ in the definition of $v_E(e_r)$, then the updated probability would equal 0.

[8] Of course, by processing the examples one at a time we abandon any pretense that the algorithm is Bayesian. In this respect, we are proceeding in the spirit of connectionist learning algorithms in which it is assumed that the sequential processing of examples in random order, as if they were actually independent, approximates the collective effect of the examples.

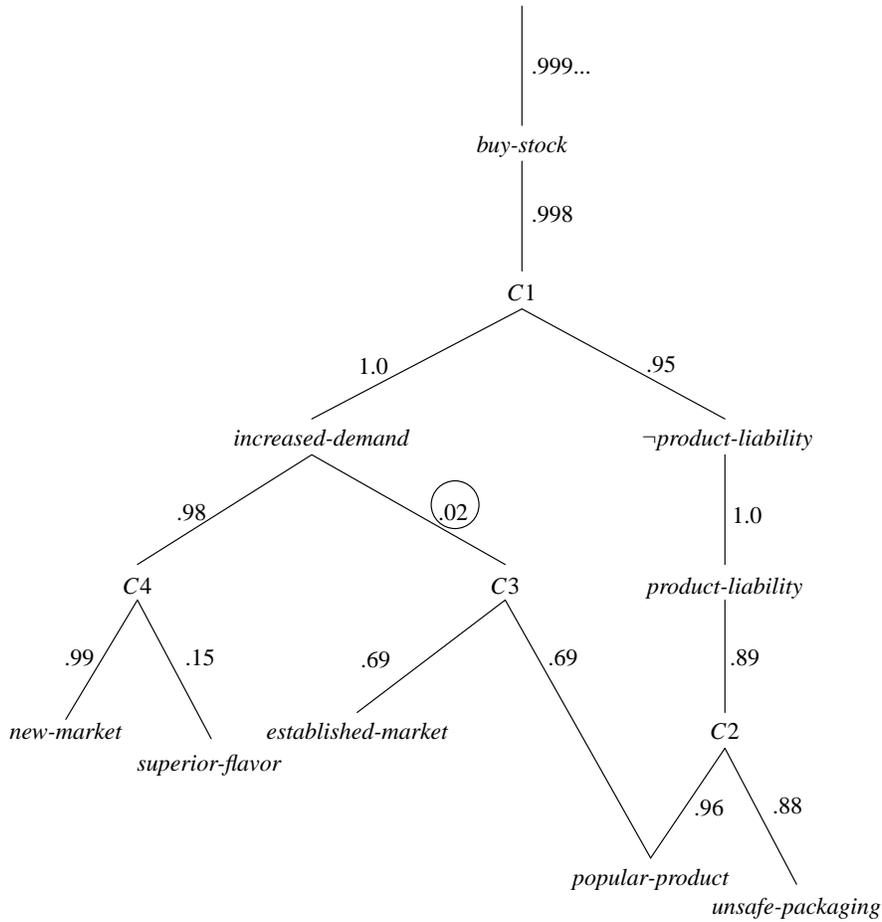

Figure 5: The weighted dt-graph of Figure 2 after processing exemplars
⟨ {*unsafe-packaging*, *new-market*}, ⟨ 1 ⟩ ⟩,
⟨ {*popular-product*, *established-market*, *superior-flavor*}, ⟨ 0 ⟩ ⟩, and
⟨ {*popular-product*, *established-market*, *celebrity-endorsement*}, ⟨ 0 ⟩ ⟩.
The clause $C3$ has dropped below the threshold.

## 4. Revising a Flawed Edge

Once an edge has been selected for revision, we must decide how to revise it. Recall that $p(e)$ represents the product of $w(e)$ and $w(n_e)$. Thus, the drop in $p(e)$ indicates either that $e$ needs to be deleted or that, less dramatically, a subtree needs to be appended to the node $n_e$. Thus, we need to determine whether to delete an edge completely or to simply weaken it by adding children; intuitively, adding edges to a clause node weakens the clause by adding conditions to its body, while adding edges to a proposition node weakens the proposition's refutation power by adding clauses to its definition. Further, if we decide to add children, then we need to determine which children to add.

## 4.1. Finding Relevant Exemplars

The first stage in making such a determination consists of establishing, for each exemplar, the role of the edge in enabling or preventing a derivation of a root. More specifically, for an IN exemplar, $\langle E, \Theta(E) \rangle$, of some root, $r$, an edge $e$ might play a positive role by facilitating a proof of $r$, or play a destructive role by preventing a proof of $r$, or may simply be irrelevant to a proof of $r$.

Once the sets of exemplars for which $e$ plays a positive role or a destructive role are determined, it is possible to append to $e$ an appropriate subtree which effectively redefines the role of $e$ such that it is used only for those exemplars for which it plays a positive role.[9] How, then, can we measure the role of $e$ in allowing or preventing a proof of $r$ from $E$?

At first glance, it would appear that it is sufficient to compare the graph $\Delta$ with the graph $\Delta_{\bar{e}}$ which results from deleting $e$ from $\Delta$. If $E \mathrel{|\!\!\!-}_\Delta r$ and $E \mathrel{|\!\!\not\!-}_{\Delta_e} r$ (or vice versa) then it is clear that $e$ is "responsible" for $r$ being provable or not provable given the exemplar $\langle E, \Theta(E) \rangle$. But, this criterion is too rigid. In the case of an OUT exemplar, even if it is the case that $E \mathrel{|\!\!\not\!-}_{\Delta_e} r$, it is still necessary to modify $e$ in the event that $e$ allowed an *additional* proof of $r$ from $E$. And, in the case of an IN exemplar, even if it is the case that $E \mathrel{|\!\!\!-}_\Delta r$ it is still necessary *not* to modify $e$ in such a way as to further prevent a proof of $r$ from $E$, since ultimately some proof is needed.

Fortunately, the weights assigned to the edges allow us the flexibility to not merely determine whether or not there is a proof of $r$ from $E$ given $\Delta$ or $\Delta_{\bar{e}}$ but also to measure numerically the flow of $E$ through $r$ both with and without $e$. This is just what is needed to design a simple heuristic which captures the degree to which $e$ contributes to a proof of $r$ from $E$.

Let $K = \langle \Delta, p \rangle$ be the weighted dt-graph which is being revised. Let $K_e = \langle \Delta, p' \rangle$ where $p'$ is identical with $p$, except that $p'(e) = 1$. Let $K_{\bar{e}} = \langle \Delta, p' \rangle$ where $p'$ is identical with $p$, except that $p'(e) = 0$; that is, $K_{\bar{e}}$ is obtained from K by deleting the edge $e$.

Then define for each root $r_i$

$$R_i(\langle E, \Theta(E) \rangle, e, K) = \frac{\left[ 1 - \Theta_i(E) \right] - u_E^{K_e}(e_{r_i})}{\left[ 1 - \Theta_i(E) \right] - u_E^{K_{\bar{e}}}(e_{r_i})}.$$

Then if $R_i(\langle E, \Theta(E) \rangle, e, K) > 2$, we say that $e$ is *needed* for $E$ and $r_i$ and if $R_i(\langle E, \Theta(E) \rangle, e, K) < 1/2$ we say that $e$ is *destructive* for $E$ and $r_i$.

---

[9] PTR is not strictly incremental in the sense that when an edge is revised its role in proving or refuting *each* exemplar is checked. If strict incrementality is a desideratum, PTR can be slightly modified so that an edge is revised on the basis of only those exemplars which have already been processed. Moreover, it is generally not necessary to check all exemplars for relevance. For example, if $e$ is an odd edge and $E$ is a correctly classified IN exemplar, then $e$ can be neither needed for $E$ (since odd edges can only make derivations more difficult) nor destructive for $E$ (since $E$ is correctly classified despite $e$).

Intuitively, this means, for example, that the edge $e$ is needed for an IN exemplar, $E$, of $r_i$, if most of the derivation of $r_i$ from $E$ passes through the edge $e$. We have simply given formal definition to the notion that "most" of the derivation passes through $e$, namely, that the flow, $u_E^{K_e}(e_{r_i})$, of $E$ through $r_i$ *without* $e$ is less than half of the flow, $u_E^{K_e}(e_{r_i})$, of $E$ through $r_i$ *with* $e$. For negation-free theories, this corresponds to the case where the edge $e$ represents a clause which is critical for the derivation of $r_i$ from $E$. The intuition for destructive edges and for OUT exemplars is analogous. Figure 6 gives the pseudo code for computing the needed and destructive sets for a given edge $e$ and exemplar set Z.

In order to understand this better, let us now return to our example dt-graph in the state in which we left it in Figure 5. The edge corresponding to the clause $C3$ has dropped below the threshold. Now let us check for which exemplars that edge is needed and for which it is destructive. Computing $R(\langle E, \Theta(E) \rangle, C3, H)$ for each example $E$ we obtain the following:

$R(\langle \{popular\text{-}product, unsafe\text{-}packaging, established\text{-}market\}, \langle 0 \rangle \rangle, C3, H) = 0.8$

$R(\langle \{unsafe\text{-}packaging, new\text{-}market\}, \langle 1 \rangle \rangle, C3, H) = 1.0$

$R(\langle \{popular\text{-}product, established\text{-}market, celebrity\text{-}endorsement\}, \langle 1 \rangle \rangle, C3, H) = 136.1$

$R(\langle \{popular\text{-}product, established\text{-}market, superior\text{-}flavor\}, \langle 0 \rangle \rangle, C3, H) = 0.1$

$R(\langle \{popular\text{-}product, established\text{-}market, ecologically\text{-}correct\}, \langle 0 \rangle \rangle, C3, H) = 0.1$

$R(\langle \{new\text{-}market, celebrity\text{-}endorsement\}, \langle 1 \rangle \rangle, C3, H) = 1.0$

**function** *Relevance*$(\langle \Delta, p \rangle : weighted\ dt\text{-}graph$, Z: *exemplar set*, $e$: *edge*): *tuple of set*;
  **begin**
    $N \Leftarrow \varnothing$;
    $D \Leftarrow \varnothing$;
    $p_{saved} \Leftarrow Copy(p)$;
    **for** $E \in$ Z **do**
      **for** $r_i \in Roots(\Delta)$ **do**
        $p(e) \Leftarrow 1; u \Leftarrow BottomUp(\Delta, p, E); u_E^{K_e} \Leftarrow u(r_i); p \Leftarrow p_{saved}$;
        $p(e) \Leftarrow 0; u \Leftarrow BottomUp(\Delta, p, E); u_E^{K_{\bar{e}}} \Leftarrow u(r_i); p \Leftarrow p_{saved}$;

        **if** $\Gamma_i(E) = 1$ **then** $R_i \Leftarrow \dfrac{u_E^{K_e}}{u_E^{K_{\bar{e}}}}$ ;

        **else** $R_i \Leftarrow \dfrac{1 - u_E^{K_e}}{1 - u_E^{K_{\bar{e}}}}$ ;
        **if** $R_i > 2$ **then** $N \Leftarrow N \cup \{E\}$;
        **if** $R_i < \dfrac{1}{2}$ **then** $D \Leftarrow D \cup \{E\}$;
      **end**
    **end**
    **return** $\langle N, D \rangle$;
  **end**

Figure 6: Pseudo code for computing the relevant sets (i.e., the needed and destructive sets) for a given edge $e$ and exemplar set Z. The general idea is to compare proof "flow" (computed using function *BottomUp*) both with and without the edge in question for each exemplar in the exemplar set. Note that the original weights are saved and later restored at the end of the computation.

The high value of

$$R(\langle\,\{popular\text{-}product, established\text{-}market, celebrity\text{-}endorsement\},\langle\,1\,\rangle\,\rangle, C3, H)$$

reflects the fact that without the clause $C3$, there is scant hope of a derivation of *buy-stock* for this example. (Of course, in principle, both *new-market* and *superior-flavor* might still be deleted from the body of clause $C4$, thus obviating the need for $C3$, but the high weight associated with the literal *new-market* in $C4$ indicates that this is unlikely.) The low values of

$$R(\langle\,\{popular\text{-}product, established\text{-}market, superior\text{-}flavor\},\langle\,0\,\rangle\,\rangle, C3, H) \text{ and}$$
$$R(\langle\,\{popular\text{-}product, established\text{-}market, ecologically\text{-}correct\},\langle\,0\,\rangle\,\rangle, C3, H)$$

reflect the fact that eliminating the clause $C3$ would greatly diminish the currently undesirably high flow through *buy-stock* (i.e., probability of a derivation of *buy-stock*) from each of these examples.

An interesting case to examine is that of

$$\langle\,\{popular\text{-}product, unsafe\text{-}packaging, established\text{-}market\},\langle\,0\,\rangle\,\rangle.$$

It is true that the elimination of $C3$ is helpful in preventing an unwanted derivation of *buy-stock* because it prevents a derivation of *increased-demand* which is necessary for *buy-stock* in clause $C1$. Nevertheless, $R$ correctly reflects the fact that the clause $C3$ is *not* destructive for this exemplar since even in the presence of $C3$, *buy-stock* is not derivable due to the failure of the literal ¬*product-liability*.

## 4.2. Appending a Subtree

Let $N$ be the set of examples for which $e$ is needed for some root and let $D$ be the set of examples for which $e$ is destructive for some root (and not needed for any other root). Having found the sets $N$ and $D$, how do we repair $e$?

At this point, if the set $D$ is non-empty and the set $N$ is empty, we simply delete the edge from $\Delta_\Gamma$. We justify this deletion by noting that no exemplars require $e$, so deletion will not compromise the performance of the theory. On the other hand, if $N$ is not empty, we apply some inductive algorithm[10] to produce a disjunctive normal form (DNF) logical expression constructed from observable propositions which is true for each exemplar in $D$ but no exemplar in $N$. We reformulate this DNF expression as a conjunction of clauses by taking a single new literal $l$ as the head of each clause, and using each conjunct in the DNF expression as the body of one of the clauses. This set of clauses is converted into dt-graph $\Delta_n$ with $l$ as its root. We then suture $\Delta_n$ to $e$ by adding to $\Delta_\Gamma$ a new node $t$, an edge from $e$ to $t$, and another edge from $t$ to the root, $l$, of $\Gamma_n$.

In order to understand why this works, first note the important fact that (like every other subroutine of PTR), this method is essentially identical whether the edge, $e$, being repaired is a clause edge, literal edge or negation edge. However, when translating back from dt-graph form to domain theory form, the new node $t$ will be interpreted differently depending on whether $n_e$ is a clause or a literal. If $n_e$ is a literal, then $t$ is interpreted as the clause $n_e \leftarrow l$. If $n_e$ is a clause,

---

[10] Any standard algorithm for constructing decision trees from positive and negative examples can be used. Our implementation of PTR uses ID3 (Quinlan, 1986). The use of an inductive component to add new substructure is due to Ourston and Mooney (Ourston & Mooney, in press).

then $t$ is interpreted as the negative literal $\neg l$.[11]

Now it is plain that those exemplars for which $e$ is destructive will use the graph rooted at $t$ to overcome the effect of $e$. If $n_e$ is a literal which undesirably excludes $E$, then $E$ will get by $n_e$ by satisfying the clause $t$; if $n_e$ is a clause which undesirably allows $E$, then $E$ will be stopped by the

---

**function** *Revise*($\langle \Delta, p \rangle$ : *weighted dt-graph* , Z : *set of exemplars*, $e$ : *edge*, $\lambda$ : *real*) : *weighted dt-graph*;
  **begin**
    $\langle N, D \rangle \Leftarrow Relevance(\langle \Delta, p \rangle, Z, e)$;
    **if** $D \neq \varnothing$ **then**
      **begin**
        **if** $N = \varnothing$ **then** $p(e) \Leftarrow 0$;
        **else**
          **begin**
            $p(e) \Leftarrow \lambda$;
            $l \Leftarrow NewLiteral()$;
            $\Delta_{ID3} = DTGraph(l, DNF\text{-}ID3(D, N))$;
            $t \Leftarrow NewNode()$; $\Delta \Leftarrow AddNode(\Delta, t)$;
            **if** *Clause?*($n_e$) **then** *Label*$(t) \Leftarrow \neg l$;
            **else** *Label*$(t) \Leftarrow NewClause()$;
            $\Delta \Leftarrow AddEdge(\Delta, \langle n_e, t \rangle)$; $p(\langle n_e, t \rangle) \Leftarrow \lambda$;
            $\Delta \Leftarrow AddEdge(\Delta, \langle t, Root(\Delta_{ID3}) \rangle)$; $p(\langle t, Root(\Delta_{ID3}) \rangle) \Leftarrow 1$;
            $\Delta \Leftarrow \Delta \bigcup \Delta_{ID3}$; for $e \in \Delta_{ID3}$ **do** $p(e) \Leftarrow 1$;
          **end**
      **end**
    **return** $\langle \Delta, p \rangle$;
  **end**

Figure 7: Pseudo code for performing a revision. The function *Revise* takes a dt-graph, a set of exemplars Z, an edge to be revised $e$, and a parameter $\lambda$ as inputs and produces a revised dt-graph as output. The function *DNF-ID3* is an inductive learning algorithm that produces a DNF formula that accepts elements of $D$ but not of $N$, while the function *DTGraph* produces a dt-graph with the given root from the given DNF expression as described in the text. For the sake of expository simplicity, we have not shown the special cases in which $n_e$ is a leaf or $e$ is a negation edge, as discussed in Footnote 11.

---

[11] Of course, if we were willing to sacrifice some elegance, we could allow separate sub-routines for the clause case and the literal case. This would allow us to make the dt-graphs to be sutured considerably more compact. In particular, if $n_e$ is a literal we could suture the children of $l$ in $\Delta_n$ directly to $n_e$. If $n_e$ is a clause, we could use the inductive algorithm to find a DNF expression which excludes examples in $D$ and includes those in $N$ (rather than the other way around as we now do it). Translating this expression to a dt-graph $\Delta$ with root $l$, we could suture $\Delta_n$ to $\Delta_\Gamma$ by simply adding an edge from the clause $n_e$ to the root $l$. Moreover, if $\Delta_n$ represents a single clause $l \leftarrow l_1, \cdots, l_m$ then we can simply suture each of the leaf-nodes $l_1, \cdots, l_m$ directly to $n_e$. Note that if $n_e$ is a leaf or a negative literal, it is inappropriate to append child edges to $n_e$. In such cases, we simply replace $n_e$ with a new literal $l'$ and append to $l'$ both $\Delta_n$ and the graph of the clause $l' \leftarrow n_e$.

new literal $t = \neg l$.

Whenever a graph $\Delta_n$ is sutured into $\Delta_\Gamma$, we must assign weights to the edges of $\Delta_n$. Unlike the original domain theory, however, the new substructure is really just an artifact of the inductive algorithm used and the current relevant exemplar set. For this reason, it is almost certainly inadvisable to try to revise it as new exemplars are encountered. Instead, we would prefer that this new structure be removed and replaced with a more appropriate new construct should the need arise. To ensure replacement instead of revision, we assign unit certainty factors to all edges of the substructure. Since the internal edges of the new structure have weights equal to 1, they will never be revised. Finally, we assign a default weight $\lambda$ to the substructure root edge $\langle n_e, t \rangle$, that connects the new component to the existing $\Delta_\Gamma$ and we reset the weight of the revised edge, $e$, to the same value $\lambda$. Figure 7 gives the pseudo code for performing the revision step just described.

Consider our example from above. We are repairing the clause $C3$. We have already found that the set $D$ consists of the examples

{*popular-product*, *established-market*, *superior-flavor*} and
{*popular-product*, *established-market*, *ecologically-correct*}

while the set $N$ consists of the single example

{*popular-product*, *established-market*, *celebrity-endorsement*}.

Using ID3 to find a formula which excludes $N$ and includes $D$, we obtain { ¬*celebrity-endorsement*} which translates into the single clause, {$l \leftarrow \neg$*celebrity-endorsement*}. Translating into dt-graph form and suturing (and simplifying using the technique of Footnote 11), we obtain the dt-graph shown in Figure 8.

Observe now that the domain theory $T'$ represented by this dt-graph correctly classifies the examples

{*popular-product*, *established-market*, *superior-flavor*} and
{*popular-product*, *established-market*, *ecologically-correct*}

which were misclassified by the original domain theory T.

## 5. The PTR Algorithm

In this section we give the details of the control algorithm which puts the pieces of the previous two sections together and determines termination.

### 5.1. Control

The PTR algorithm is shown in Figure 9. We can briefly summarize its operation as follows:

(1)   PTR process exemplars in random order, updating weights and performing revisions when necessary.

(2)   Whenever a revision is made, the domain theory which corresponds to the newly revised graph is checked against all exemplars.

(3)   PTR terminates if:
      (i)   All exemplars are correctly classified, or
      (ii)  Every edge in the newly revised graph has weight 1.

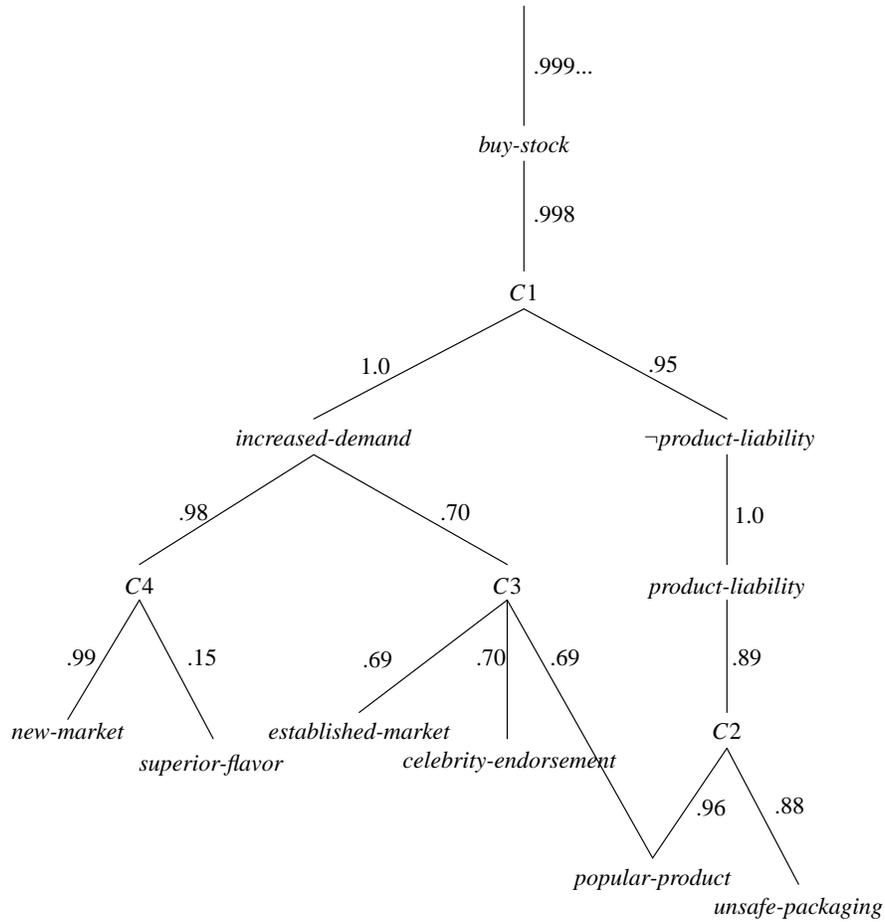

Figure 8: The weighted dt-graph of Figure 2 after revising the clause $C3$ (the graph has been slightly simplified in accordance with the remark in Footnote 11).

(4)  If, after a revision is made, PTR does not terminate, then it continues processing exemplars in random order.

(5)  if, after a complete cycle of exemplars has been processed, there remain misclassified exemplars, then we
  (i)  Increment the revision threshold $\sigma$ so that $\sigma = \min[\sigma + \delta_\sigma, 1]$, and
  (ii) Increment the value $\lambda$ assigned to a revised edge and to the root edge of an added component, so that $\lambda = \min[\lambda + \delta_\lambda, 1]$.

(6)  Now we begin anew, processing the exemplars in (new) random order.

It is easy to see that PTR is guaranteed to terminate. The argument is as follows. Within $\max\left[\dfrac{1}{\delta_\sigma}, \dfrac{1}{\delta_\lambda}\right]$ cycles, both $\sigma$ and $\lambda$ will reach 1. At this point, every edge with weight less than 1 will be revised and will either be deleted or have its weight reset to $\lambda = 1$. Moreover, any edges added during revision will also be assigned certainty factor $\lambda = 1$. Thus all edges will have weight 1 and the algorithm terminates by the termination criterion (ii).

Now, we wish to show that PTR not only terminates, but that it terminates with every exemplar correctly classified. That is, we wish to show that, in fact, termination criterion (ii) can never be satisfied unless termination criterion (i) is satisfied as well. We call this property *convergence*. In Appendix C we prove that, under certain very general conditions, PTR is guaranteed to converge.

## 5.2. A Complete Example

Let us now review the example which we have been considering throughout this paper.

We begin with the flawed domain theory and set of exemplars introduced in Section 1.

$C1$: *buy-stock* ← *increased-demand* ∧ ¬*product-liability*
$C2$: *product-liability* ← *popular-product* ∧ *unsafe-packaging*
$C3$: *increased-demand* ← *popular-product* ∧ *established-market*
$C4$: *increased-demand* ← *new-market* ∧ *superior-flavor*.

We translate the domain theory into the weighted dt-graph $\langle \Delta_\mathrm{T}, p \rangle$ of Figure 2, assigning weights via a combination of user-provided information and default values. For example, the user has indicated that their confidence in the first literal (*increased-demand*) in the body of clause $C1$ is greater than their confidence in the second literal ( ¬*product-liability*).

> **function** *PTR*($\langle \Delta, p \rangle$: *weighted dt-graph*, Z: *set of exemplars*,
>  $\langle \lambda_0, \sigma_0, \delta_\lambda, \delta_\sigma, \varepsilon \rangle$: *five tuple of real*): *weighted dt-graph*;
> **begin**
>  $\lambda \Leftarrow \lambda_0$;
>  $\sigma \Leftarrow \sigma_0$;
>  **while** $\exists E \in \mathrm{Z}$ **such that** $\Gamma(E) \neq \Theta(E)$ **do**
>   **begin**
>    **for** $E \in RandomlyPermute(\mathrm{Z})$ **do**
>     **begin**
>      $u \Leftarrow BottomUp(\langle \Delta, p \rangle, E)$;
>      $\langle \Delta, p \rangle \Leftarrow TopDown(\langle \Delta, p \rangle, u, E, \varepsilon)$;
>      **if** $\exists e \in \Delta$ **such that** $p(e) \leq \sigma$ **then** $\langle \Delta, p \rangle \Leftarrow Revise(\langle \Delta, p \rangle, \mathrm{Z}, \varepsilon, \lambda)$;
>      **if** $\forall e \in \Delta, p(e) = 1$ **or** $\forall E \in \mathrm{Z}, \Gamma(E) = \Theta(E)$ **then return** $\langle \Delta, p \rangle$;
>     **end**
>    $\lambda \Leftarrow \max[\lambda + \delta_\lambda, 1]$;
>    $\sigma \Leftarrow \max[\sigma + \delta_\sigma, 1]$;
>   **end**
> **end**

Figure 9: The PTR control algorithm. Input to the algorithm consists of a weighted dt-graph $\langle \Delta, p \rangle$, a set of exemplars Z, and five real-valued parameters $\lambda_0, \sigma_0, \delta_\lambda, \delta_\sigma$, and $\varepsilon$. The algorithm produces a revised weighted dt-graph whose implicit theory correctly classifies all exemplars in Z.

We set the revision threshold $\sigma$ to .1, the reset value $\lambda$ initially to .7 and their respective increments $\delta_\sigma$ and $\delta_\lambda$ to .03. We now start updating the weights of the edges by processing the exemplars in some random order.

We first process the exemplar

$\langle\{$*unsafe-packaging, new-market*$\}, \langle 1 \rangle\rangle$.

First, the leaves of the dt-graph are labeled according to their presence or absence in the exemplar. Second, $u_E(e)$ values (proof flow) are computed for all edges of the dt-graph in bottom up fashion. Next, $v_E(e_{r_i})$ values are set to reflect the vector of correct classifications for the example $\Theta(E)$. New values for $v_E(e)$ are computed in top down fashion for each edge in the dt-graph. As these values are computed, new values for $p(e)$ are also computed. Processing of this first exemplar produces the updated dt-graph shown in Figure 3.

Processing of exemplars continues until either an edge weight falls below $\sigma$ (indicating a flawed domain theory element has been located), a cycle (processing of all known exemplars) is completed, or the PTR termination conditions are met. For our example, after processing the additional exemplars

$\langle\{$*popular-product, established-market, superior-flavor*$\}, \langle 0 \rangle\rangle$ and
$\langle\{$*popular-product, established-market, ecologically-correct*$\}, \langle 0 \rangle\rangle$

the weight of the edge corresponding to clause $C3$ drops below $\sigma$ (see Figure 5), indicating that this edge needs to be revised.

We proceed with the revision by using the heuristic in Section 4.2 in order to determine for which set of exemplars the edge in question is needed and for which it is destructive. The edge corresponding to the clause $C3$ is needed for

$\{\langle\{$*popular-product, established-market, celebrity-endorsement*$\}, \langle 1 \rangle\rangle\}$

and is destructive for

$\{\langle\{$*popular-product, established-market, ecologically-correct*$\}, \langle 0 \rangle\rangle,$
$\langle\{$*popular-product, established-market, superior-flavor*$\}, \langle 0 \rangle\rangle\}$.

Since the set for which the edge is needed is not empty, PTR chooses to append a subtree weakening clause $C3$ rather than simply deleting the clause outright. Using these sets as input to ID3, we determine that the fact *celebrity-endorsement* suitably discriminates between the needed and destructive sets. We then repair the graph to obtain the weighted dt-graph shown in Figure 8. This graph corresponds to the theory in which the literal *celebrity-endorsement* has been added to the body of $C3$.

We now check the newly-obtained theory embodied in the dt-graph of Figure 8 (i.e., ignoring weights) against all the exemplars and determine that there are still misclassified exemplars, namely

$\langle\{$*unsafe-packaging, new-market*$\}, \langle 1 \rangle\rangle$ and
$\langle\{$*new-market, celebrity-endorsement*$\}, \langle 1 \rangle\rangle$.

Thus, we continue processing the remaining exemplars in the original (random) order.

After processing the exemplars

$\langle\{$*popular-product, unsafe-packaging, established-market*$\}, \langle 0 \rangle\rangle,$
$\langle\{$*popular-product, established-market, celebrity-endorsement*$\}, \langle 1 \rangle\rangle$, and
$\langle\{$*new-market, celebrity-endorsement*$\}, \langle 1 \rangle\rangle,$

the weight of the edge corresponding to the literal *superior-flavor* in clause *C*4 drops below the revision threshold $\sigma$. We then determine that this edge is not needed for any exemplar and thus the edge is simply deleted.

At this point, no misclassified exemplars remain. The final domain theory is:

*C*1: *buy-stock* $\leftarrow$ *increased-demand* $\wedge$ $\neg$*product-liability*
*C*2: *product-liability* $\leftarrow$ *popular-product* $\wedge$ *unsafe-packaging*
*C*3: *increased-demand* $\leftarrow$ *popular-product* $\wedge$ *established-market* $\wedge$ *celebrity-endorsement*
*C*4: *increased-demand* $\leftarrow$ *new-market*.

This theory correctly classifies all known exemplars and PTR terminates.

## 6. Experimental Evaluation

In this section we will examine experimental evidence that illustrates several fundamental hypotheses concerning PTR. We wish to show that:

(1) theories produced by PTR are of high quality in three respects: they are of low radicality, they are of reasonable size, and they provide accurate information regarding exemplars other than those used in the training.

(2) PTR converges rapidly — that is, it requires few cycles to find an adequate set of revisions.

(3) well-chosen initial weights provided by a domain expert can significantly improve the performance of PTR.

More precisely, given a theory $\Gamma'$ obtained by using PTR to revise a theory $\Gamma$ on the basis of a set of training examplars, we will test these hypotheses as follows.

*Radicality*. Our claim is that $Rad_{\mathrm{K}}(\Gamma')$ is typically close to minimal over all theories which correctly classify all the examples. For cases where the target theory, $\Theta$, is known, we measure $\dfrac{Rad_{\mathrm{K}}(\Gamma')}{Rad_{\mathrm{K}}(\Theta)}$. If this value is less than 1, then PTR can be said to have done even "better" than finding the target theory in the sense that it was able to correctly classify all training examples using less radical revisions than those required to restore the target theory. If the value is greater than 1, then PTR can be said to have "over-revised" the theory.

*Cross-validation*. We perform one hundred repetitions of cross-validation using nested sets of training examples. It should be noted that our actual objective is to minimize radicality, and that often there are theories that are less radical than the target theory which also satisfy all training examples. Thus, while cross-validation gives some indication that theory revision is being successfully performed, it is not a primary objective of theory revision.

*Theory size*. We count the number of clauses and literals in the revised theory merely to demonstrate that theories obtained using PTR are comprehensible. Of course, the precise size of the theory obtained by PTR is largely an artifact of the choice of inductive component.

*Complexity*. Processing a complete cycle of exemplars is $O(n \times d)$ where $n$ is the number of edges in the graph and $d$ is the number of exemplars. Likewise repairing an edge is $O(n \times d)$. We will measure the number of cycles and the number of repairs made until convergence. (Recall that the number of cycles until convergence is in any event bounded by $\max \left[ \dfrac{1}{\delta_\sigma}, \dfrac{1}{\delta_\lambda} \right]$. We will show that, in practice, the number of cycles is small even if $\delta_\sigma = \delta_\lambda = 0$.

*Utility of Bias.* We wish to show that user-provided guidance in choosing initial weights leads to faster and more accurate results. For cases in which the target theory, $\Theta$, is known, let $S$ be the set of edges of $\Delta_\Gamma$ which need to be revised in order to restore the target theory $\Theta$. Define $p_\beta(e)$ such that for each $e \in S$, $1 - p_\beta(e) = (1 - p(e))^{\frac{1}{\beta}}$ and for each $e \notin S$, $p_\beta(e) = (p(e))^{\frac{1}{\beta}}$. That is, each edge which needs to be revised to obtain the intended theory has its initial weight diminished and each edge which need not be revised to obtain the intended theory has its weight increased. Let $K_\beta = \langle \Delta_\Gamma, p_\beta \rangle$. Then, for each $\beta$,

$$Rad_{K_\beta}(\Theta) = -\log(\prod_{e \in S}(1 - p(e))^{\frac{1}{\beta}} \times \prod_{e \notin S}(p(e))^{\frac{1}{\beta}}) = \frac{1}{\beta} Rad_K(\Theta).$$

Here, we compare the results of cross-validation and number-of-cycles experiments for $\beta = 2$ with their unbiased counterparts (i.e., $\beta = 1$).

## 6.1. Comparison with other Methods

In order to put our results in perspective we compare them with results obtained by other methods.[12]

(1) ID3 (Quinlan, 1986) is the inductive component we use in PTR. Thus using ID3 is equivalent to learning directly from the examples without using the initial flawed domain theory. By comparing results obtained using ID3 with those obtained using PTR we can gauge the usefulness of the given theory.

(2) EITHER (Ourston & Mooney, in press) uses enumeration of partial proofs in order to find a minimal set of literals, the repair of which will satisfy all the exemplars. Repairs are then made using an inductive component. EITHER is exponential in the size of the theory. It cannot handle theories with negated internal literals. It also cannot handle theories with multiple roots unless those roots are mutually exclusive.

(3) KBANN (Towell & Shavlik, 1993) translates a symbolic domain theory into a neural net, uses backpropagation to adjust the weights of the net's edges, and then translates back from net form to partially symbolic form. Some of the rules in the theory output by KBANN might be numerical, i.e., not strictly symbolic.

(4) RAPTURE (Mahoney & Mooney, 1993) uses a variant of backpropagation to adjust certainty factors in a probabilistic domain theory. If necessary, it can also add a clause to a root. All the rules produced by RAPTURE are numerical. Like EITHER, RAPTURE cannot handle negated internal literals or multiple roots which are not mutually exclusive.

Observe that, relative to the other methods considered here, PTR is liberal in terms of the theories it can handle, in that (like KBANN, but unlike EITHER and RAPTURE) it can handle negated literals and non-mutually exclusive multiple roots; it is also strict in terms of the theories it yields in that (like EITHER, but unlike KBANN and RAPTURE) it produces strictly symbolic theories.

---

[12] There are other interesting theory revision algorithms, such as RTLS (Ginsberg, 1990), for which no comparable data is available.

We have noted that both KBANN and RAPTURE output "numerical" rules. In the case of KBANN, a numerical rule is one which fires if the sum of weights associated with satisfied antecedents exceeds a threshold. In the case of RAPTURE, the rules are probabilistic rules using certainty factors along the lines of MYCIN (Buchanan & Shortliffe, 1984). One might ask, then, to what extent are results obtained by theory revision algorithms which output numerical rules merely artifacts of the use of such numerical rules? In other words, can we separate the effects of using numerical rules from the effects of learning?

To make this more concrete, consider the following simple method for transforming a symbolic domain theory into a probabilistic domain theory and then reclassifying examples using the obtained probabilistic theory. Suppose we are given some possibly-flawed domain theory $\Gamma$. Suppose further that we are not given the classification of even a single example. Assign a weight $p(e)$ to each edge of $\Delta_\Gamma$ according to the default scheme of Appendix A. Now, using the bottom-up subroutine of the updating algorithm, compute $u_E(e_r)$ for each test example $E$. (Recall that $u_E(e_r)$ is a measure of how close to a derivation of $r$ from $E$ there is, given the weighted dt-graph $\langle \Delta_\Gamma, p \rangle$.) Now, for some chosen "cutoff" value $0 \leq n \leq 100$, if $E_0$ is such that $u_{E_0}(e_r)$ lies in the upper $n\%$ of the set of values $\{u_E(e_r)\}$ then conclude that $\Gamma$ is true for $E_0$; otherwise conclude that $\Gamma$ is false for $E_0$.

This method, which for the purpose of discussion we call PTR*, does not use any training examples at all. Thus if the results of theory revision systems that employ numerical rules can be matched by PTR* — *which performs no learning* — then it is clear that the results are merely artifacts of the use of numerical rules.

## 6.2. Results on the PROMOTER Theory

We first consider the PROMOTER theory from molecular biology (Murphy & Aha, 1992), which is of interest solely because it has been extensively studied in the theory revision literature (Towell & Shavlik, 1993), thus enabling explicit performance comparison with other algorithms. The PROMOTER theory is a flawed theory intended to recognize promoters in DNA nucleotides. The theory recognized none of a set of 106 examples as promoters despite the fact that precisely half of them are indeed promoters.[13]

Unfortunately, the PROMOTER theory (like many others used in the theory revision literature) is trivial in that it is very shallow. Moreover, it is atypical of flawed domains in that it is overly specific but not overly general. Given the shortcomings of the PROMOTER theory, we will also test PTR on a synthetically-generated theory in which errors have been artificially introduced. These synthetic theories are significantly deeper than those used to test previous methods. Moreover, the fact that the intended theory is known will enable us to perform experiments involving radicality and bias.

---

[13] In our experiments, we use the default initial weights assigned by the scheme of Appendix A. In addition, the clause whose head is the proposition *contact* is treated as a definition not subject to revision but only deletion as a whole.

### 6.2.1. Cross-validation

In Figure 10 we compare the results of cross-validation for PROMOTER. We distinguish between methods which use numerical rules (top plot) and those which are purely symbolic (bottom plot).

The lower plot in Figure 10 highlights the fact that, using the value $n = 50$, PTR* achieves better accuracy, *using no training examples*, than any of the methods considered here achieve using 90 training examples. In particular, computing $u_E(e_r)$ for each example, we obtain that of the 53 highest-ranking examples 50 are indeed promoters (and, therefore, of the 53 lowest-ranking examples 50 are indeed non-promoters). Thus, PTR* achieves $94.3\%$ accuracy. (In fact, all of the 47 highest-ranking examples are promoters and all of the 47 lowest-ranking are not promoters. Thus, a more conservative version of PTR* which classifies the, say, 40% highest-ranking examples as IN and the 40% lowest-ranking as OUT, would indeed achieve 100% accuracy over the examples for which it ventured a prediction.)

This merely shows that the original PROMOTER theory is very accurate provided that it is given a numerical interpretation. Thus we conclude that the success of RAPTURE and KBANN for this domain is not a consequence of learning from examples but rather an artifact of the use of numerical rules.

As for the three methods — EITHER, PTR and ID3 — which yield symbolic rules, we see in the top plot of Figure 10 that, as reported in (Ourston & Mooney, in press; Towell & Shavlik, 1993), the methods which exploit the given flawed theory do indeed achieve better results on PROMOTER than ID3, which does not exploit the theory. Moreover, as the size of the training set grows, the performance of PTR is increasingly better than that of EITHER.[14]

Finally, we wish to point out an interesting fact about the example set. There is a set of 13 out of the 106 examples which each contain information substantially different than that in the rest of the examples. Experiments show that using ten-fold cross-validation on the 93 "good" examples yields $99.2\%$ accuracy, while training on all 93 of these examples and testing on the 13 "bad" examples yields below 40% accuracy.

### 6.2.2. Theory size

The size of the output theory is an important measure of the comprehensibility of the output theory. Ideally, the size of the theory should not grow too rapidly as the number of training examples is increased, as larger theories are necessarily harder to interpret. This observation holds both for the number of clauses in the theory as well as for the average number of antecedents in each of those clauses.

Theory sizes for the theories produced by PTR are shown in Figure 11. The most striking aspect of these numbers is that all measures of theory size are relatively stable with respect to training set size. Naturally, the exact values are to a large degree an artifact of the inductive learning component used. In contrast, for EITHER, theory size increases with training set size (Ourston, 1991). For example, for 20 training examples the output theory size (clauses plus literals) is 78, while for 80 training examples, the output theory size is 106.

---

[14] Those readers familiar with the PROMOTER theory should note that the improvement over EITHER is a consequence of PTR repairing one flaw at a time and using a sharper relevance criterion. This results in PTR always deleting the extraneous *conformation* literal, while EITHER occasionally fails to do so, particularly as the number of training exmaple increases.

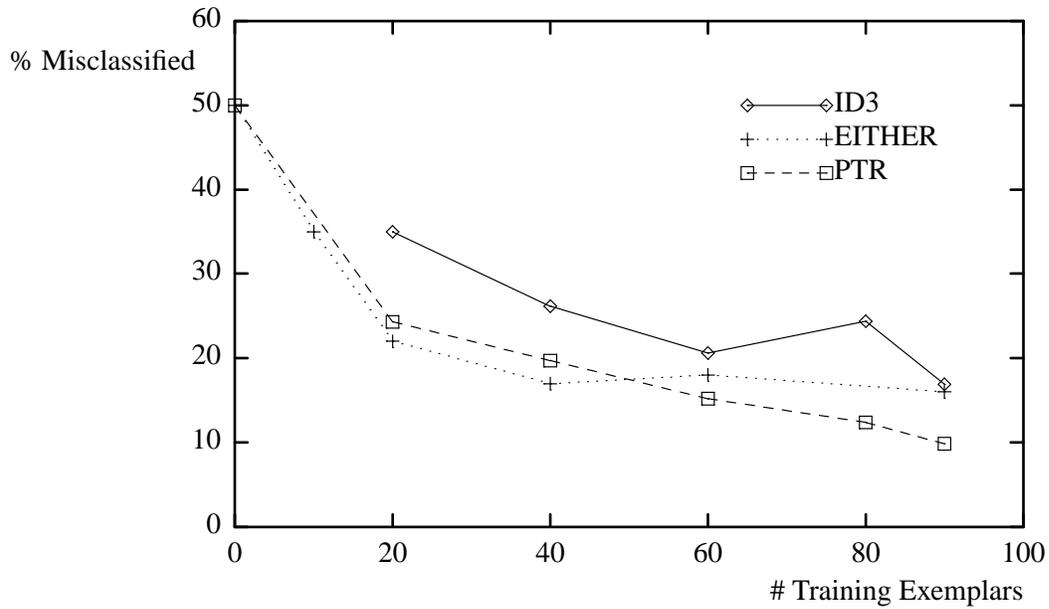

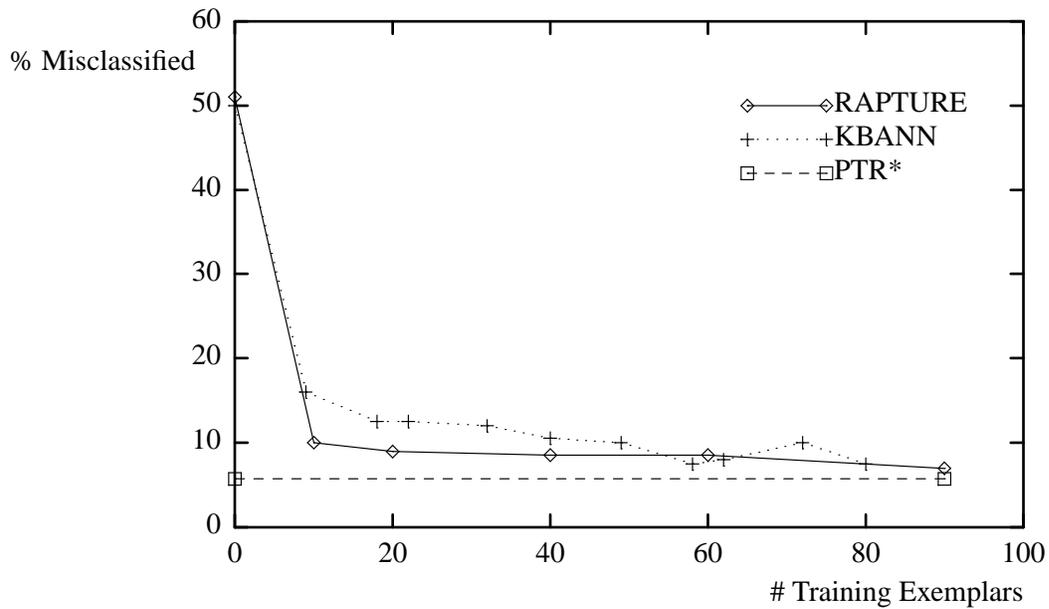

Figure 10: PROMOTER: Error rates using nested training sets for purely symbolic theories (top plot) and numeric theories (bottom plot). Results for EITHER, RAPTURE, and KBANN are taken from (Mahoney & Mooney, 1993), while results for ID3 and PTR were generated using similar experimental procedures. Recall that PTR* is a non-learning numerical rule system; the PTR* line is extended horizontally for clarity.

| Training Set Size | Mean Clauses in Output | Mean Literals in Output | Mean Revisions to Convergence | Mean Exemplars to Convergence |
|---|---|---|---|---|
| *Original Theory* | 14 | 83 | | |
| 20 | 11 | 39 | 10.7 | 88 |
| 40 | 11 | 36 | 15.2 | 140 |
| 60 | 11 | 35 | 18.2 | 186 |
| 80 | 11 | 32 | 22.1 | 232 |
| 100 | 12 | 36 | 22.0 | 236 |

Figure 11: PROMOTER: Results. Numbers reported for each training set size are average values over one hundred trials (ten trials for each of ten example partitions).

Unfortunately, making direct comparisons with KBANN or RAPTURE is difficult. In the case of KBANN and RAPTURE, which allow numerical rules, comparison is impossible given the differences in the underlying representation languages. Nevertheless, it is clear that, as expected, KBANN produces significantly larger theories than PTR. For example, using 90 training examples from the PROMOTER theory, KBANN produces numerical theories with, on average, 10 clauses and 102 literals (Towell & Shavlik, 1993). These numbers would grow substantially if the theory were converted into strictly symbolic terms. RAPTURE, on the other hand, does not change the theory size, but, like KBANN, yields numerical rules (Mahoney & Mooney, 1993).

### 6.2.3. Complexity

EITHER is exponential in the size of the theory and the number of training examples. For KBANN, each cycle of the training-by-backpropagation subroutine is $O(d \times n)$ (where $d$ is the size of the network and $n$ is the number of exemplars), and the number of such cycles typically numbers in the hundreds even for shallow nets.

Like backpropagation, the cost of processing an example with PTR is linear in the size of the theory. In contrast, however, PTR typically converges after processing only a tiny fraction of the number of examples required by standard backpropagation techniques. Figure 11 shows the average number of exemplars (not cycles!) processed by PTR until convergence as a function of training set size. The only other cost incurred by PTR is that of revising the theory. Each such revision in $O(d \times n)$. The average number of revisions to convergence is also shown in Figure 11.

### 6.3. Results on Synthetic Theories

The character of the PROMOTER theory make it less than ideal for testing theory revision algorithms. We wish to consider theories which (i) are deeper, which (ii) make substantial use of negated internal literals and which (iii) are overly general as well as overly specific. As opposed to shallow theories which can generally be easily repaired at the leaf level, deeper theories often require repairs at internal levels of the theory. Therefore, a theory revision algorithm which may perform well on shallow theories will not necessarily scale up well to larger theories. Moreover, as theory size increases, the computational complexity of an algorithm might preclude its application altogether. We wish to show that PTR scales well to larger, deeper theories.

Since deeper, propositional, real-world theories are scarce, we have generated them synthetically. As an added bonus, we now know the target theory so we can perform controlled experiments on bias and radicality. In (Feldman, 1993) the aggregate results of experiments performed on a collection of synthetic theories are reported. In order to avoid the dubious practice of averaging results over different theories and in order to highlight significant features of a particular application of PTR, we consider here one synthetic theory typical of those studied in (Feldman, 1993).

$$r \leftarrow A, B$$
$$r \leftarrow C, \neg D$$
$$A \leftarrow E, F$$
$$A \leftarrow p_0, \neg G, p_1, p_2, p_3$$
$$B \leftarrow \neg p_0$$
$$B \leftarrow p_1, \neg H$$
$$B \leftarrow p_4, \neg p_{11}$$
$$C \leftarrow I, J$$
$$C \leftarrow p_2, \neg K$$
$$C \leftarrow \neg p_8, \neg p_9$$
$$D \leftarrow p_{10}, \neg p_{12}, L$$
$$D \leftarrow p_3, \neg p_9, \neg M$$
$$E \leftarrow N, p_5, p_6$$
$$E \leftarrow \neg O, \neg p_7, \neg p_8$$
$$F \leftarrow p_4$$
$$F \leftarrow Q, \neg R$$
$$G \leftarrow S, \neg p_3, p_8$$
$$G \leftarrow \neg p_{10}, p_{12}$$
$$H \leftarrow U, V$$
$$H \leftarrow p_1, p_2; p_3, p_4$$
$$I \leftarrow W$$
$$I \leftarrow p_6$$
$$J \leftarrow X, p_5$$
$$J \leftarrow Y$$
$$K \leftarrow P, \neg p_5, p_9$$
$$K \leftarrow \neg p_6, p_9$$

$$L \leftarrow T, p_1$$
$$L \leftarrow p_2, p_{12}, p_{16}$$
$$M \leftarrow Z, \neg p_{17}$$
$$M \leftarrow p_{18}, \neg p_{19}$$
$$N \leftarrow \neg p_0, p_1$$
$$N \leftarrow p_3, p_4, p_6$$
$$N \leftarrow p_{10}, \neg p_{12}$$
$$Z \leftarrow p_2, p_3$$
$$Z \leftarrow \neg p_2, p_3, p_{17}, p_{18}, p_{20}$$
$$O \leftarrow \neg p_3, p_4, p_5, p_{11}, \neg p_{12}$$
$$O \leftarrow \neg p_{13}, p_{18}$$
$$Y \leftarrow p_4, p_5 p_6$$
$$P \leftarrow \neg p_6, p_7, p_8$$
$$X \leftarrow p_7, p_9$$
$$Q \leftarrow p_0, p_4$$
$$Q \leftarrow p_3, \neg p_{13}, p_{14}, p_{15}$$
$$W \leftarrow p_{10}, p_{11}$$
$$W \leftarrow p_3, p_9$$
$$R \leftarrow p_{12}, \neg p_{13}, p_{14}$$
$$V \leftarrow \neg p_{14}, p_{15}$$
$$S \leftarrow p_3, p_6, \neg p_{14}, p_{15}, p_{16}$$
$$U \leftarrow p_{11}, p_{12}$$
$$U \leftarrow p_{13}, p_{14}, \neg p_{15}, \neg p_{16}, \neg p_{17}$$
$$T \leftarrow p_7$$
$$T \leftarrow \neg p_7, p_8, p_9, \neg p_{16}, \neg p_{17}, \neg p_{18}$$

Figure 12: The synthetic domain theory $\Theta$ used for the experiments of Section 6.

The theory $\Theta$ is shown is Figure 12. Observe that $\Theta$ includes four levels of clauses and has many negated internal nodes. It is thus substantially deeper than theories considered before in testing theory revision algorithms. We artificially introduce, in succession, 15 errors into the theory $\Theta$. The errors are shown in Figure 13. For each of these theories, we use the default initial weights assigned by the scheme of Appendix A.

Let $\Gamma_i$ be the theory obtained after introducing the first $i$ of these errors. In Figure 14 we show the radicality, $Rad_{\Gamma_i}(\Theta)$, of $\Theta$ relative to each of the flawed theories, $\Gamma_i$ for $i = 3, 6, 9, 12, 15$, as well as the number of examples misclassified by each of those theories. Note that, in general, the number of misclassified examples cannot necessarily be assumed to increase monotonically with the number of errors introduced since introducing an error may either generalize or specialize the theory. For example, the fourth error introduced is "undone" by the fifth error. Nevertheless, it is the case that for this particular set of errors, each successive theory is more radical and misclassifies a larger number of examples with respect to $\Theta$.

To measure radicality and accuracy, we choose 200 exemplars which are classified according to $\Theta$. Now for each $\Gamma_i$ ($i = 3, 6, 9, 12, 15$), we withhold 100 test examples and train on nested sets of 20, 40, 60, 80 and 100 training examples. We choose ten such partitions and run ten trials for each partition.

In Figure 15, we graph the average value of $\dfrac{Rad_{\Gamma_i}(\Gamma')}{Rad_{\Gamma_i}(\Theta)}$, where $\Gamma'$ is the theory produced by PTR. As can be seen, this value is consistently below 1. This indicates that the revisions found by PTR are less radical than what is needed to restore the original $\Theta$. *Thus by the criterion of success that PTR set for itself, minimizing radicality, PTR does better than restoring $\Theta$.* As is to be expected, the larger the training set the closer this value is to 1. Also note that as the number of errors introduced increases, the saving in radicality achieved by PTR increases as well, since a larger number of opportunities are created for more parsimonious revision. More precisely, the average number of revisions made by PTR to $\Gamma_3, \Gamma_6, \Gamma_9, \Gamma_{12}$, and $\Gamma_{15}$ with a 100 element training set are 1.4, 4.1, 7.6, 8.3, and 10.4, respectively.

| | |
|---|---|
| 1 | Added clause $A \leftarrow \neg p_6$ |
| 2 | Added clause $S \leftarrow \neg p_5$ |
| 3 | Added clause $A \leftarrow p_8, \neg p_{15}$ |
| 4 | Added literal $\neg p_6$ to clause $B \leftarrow p_4, \neg p_{11}$ |
| 5 | Deleted clause $B \leftarrow p_4, \neg p_6, \neg p_{11}$ |
| 6 | Added clause $D \leftarrow \neg p_{14}$ |
| 7 | Added clause $G \leftarrow \neg p_{12}, p_8$ |
| 8 | Added literal $p_2$ to clause $A \leftarrow E, F$ |
| 9 | Added clause $L \leftarrow p_{16}$ |
| 10 | Added clause $M \leftarrow \neg p_{13}, \neg p_7$ |
| 11 | Deleted clause $Q \leftarrow p_3, \neg p_{13}, p_{14}, p_{15}$ |
| 12 | Deleted clause $L \leftarrow p_2, p_{12}, p_{16}$ |
| 13 | Added clause $J \leftarrow p_{11}$ |
| 14 | Deleted literal $p_4$ from clause $F \leftarrow p_4$ |
| 15 | Deleted literal $p_1$ from clause $B \leftarrow p_1, \neg H$ |

Figure 13: The errors introduced into the synthetic theory $\Theta$ in order to produce the flawed synthetic theories $\Gamma_i$. Note that the fifth randomly-generated error obviates the fourth.

| | $\Gamma_3$ | $\Gamma_6$ | $\Gamma_9$ | $\Gamma_{12}$ | $\Gamma_{15}$ |
|---|---|---|---|---|---|
| Number of Errors | 3 | 6 | 9 | 12 | 15 |
| $Rad(\Theta)$ | 7.32 | 17.53 | 22.66 | 27.15 | 33.60 |
| Misclassified IN | 0 | 26 | 34 | 34 | 27 |
| Misclassified OUT | 50 | 45 | 45 | 46 | 64 |
| Initial Accuracy | 75% | 64.5% | 60.5% | 60% | 54.5% |

Figure 14: Descriptive statistics for the flawed synthetic theories $\Gamma_i$ ($i = 3, 6, 9, 12, 15$).

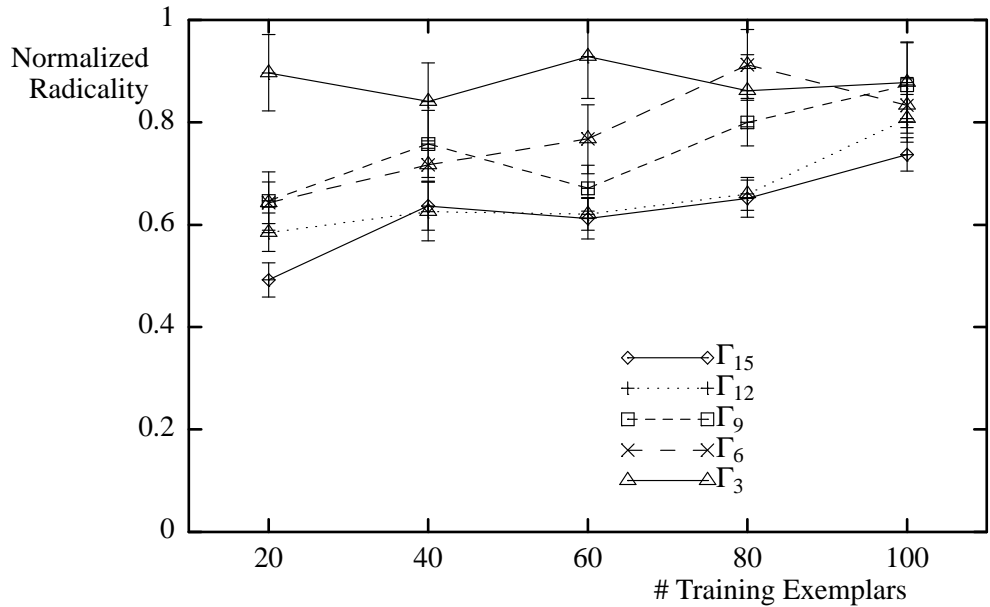

Figure 15: The normalized radicality, $\dfrac{Rad_{\Gamma_i}(\Gamma')}{Rad_{\Gamma_i}(\Theta)}$, for the output theories $\Gamma'$ produced by PTR from $\Gamma_i$ ($i = 3, 6, 9, 12, 15$). Error bars reflect 1 standard error.

An example will show how PTR achieves this. Note from Figure 13 that the errors introduced in $\Gamma_3$ are the additions of the rules:

$$A \leftarrow \neg p_6$$
$$S \leftarrow \neg p_5$$
$$S \leftarrow p_8, \neg p_{15}.$$

In most cases, PTR quickly locates the extraneous clause $A \leftarrow \neg p_6$, and discovers that deleting it results in the correct classification of all exemplars in the training set. In fact, this change also results in the correct classification of all test examples as well. The other two added rules do not affect the classification of any training examples, and therefore are not deleted or repaired by PTR. Thus the radicality of the changes made by PTR is lower than that required for restoring the original theory. In a minority of cases, PTR first deletes the clause $B \leftarrow \neg p_0$ and only then deletes the clause $A \leftarrow p_6$. Since the literal $B$ is higher in the tree than the literal $S$, the radicality of these changes is marginally higher that that required to restore the original theory.

In Figure 16, we graph the accuracy of $\Gamma'$ on the test set. As expected, accuracy degenerates somewhat as the number of errors is increased. Nevertheless, even for $\Gamma_{15}$, PTR yields theories which generalize accurately.

Figure 17 shows the average number of exemplars required for convergence. As expected, the fewer errors in the theory, the fewer exemplars PTR requires for convergence. Moreover, the number of exemplars processed grows less than linearly with the training set size. In fact, in no case was the average number of examples processed greater than 4 times the training set size. In comparison, backpropagation typically requires hundreds of cycles when it converges.

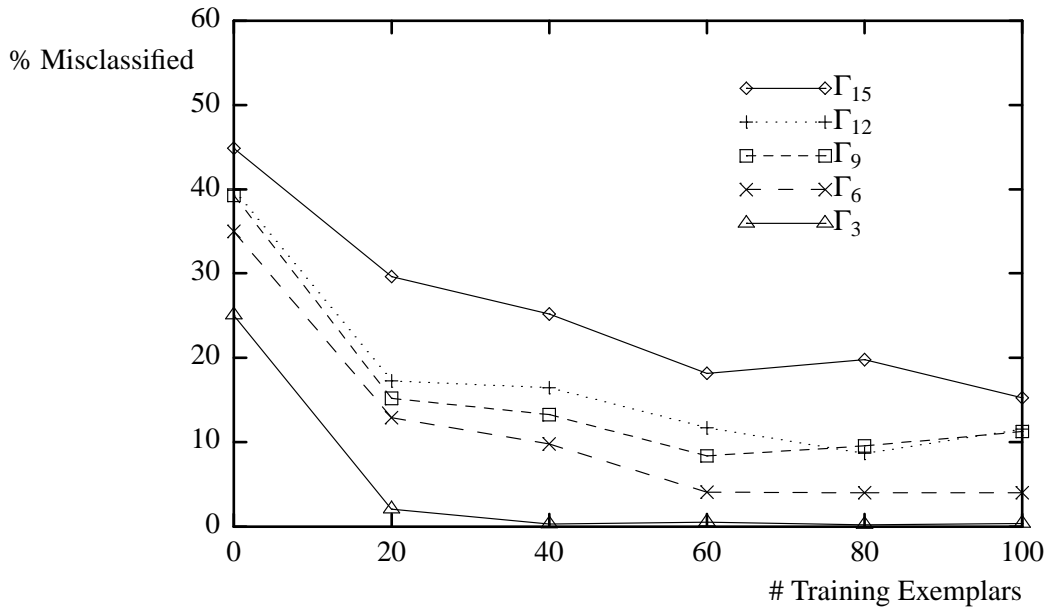

Figure 16: Error rates for the output theories produced by PTR from $\Gamma_i$ ($i = 3, 6, 9, 12, 15$).

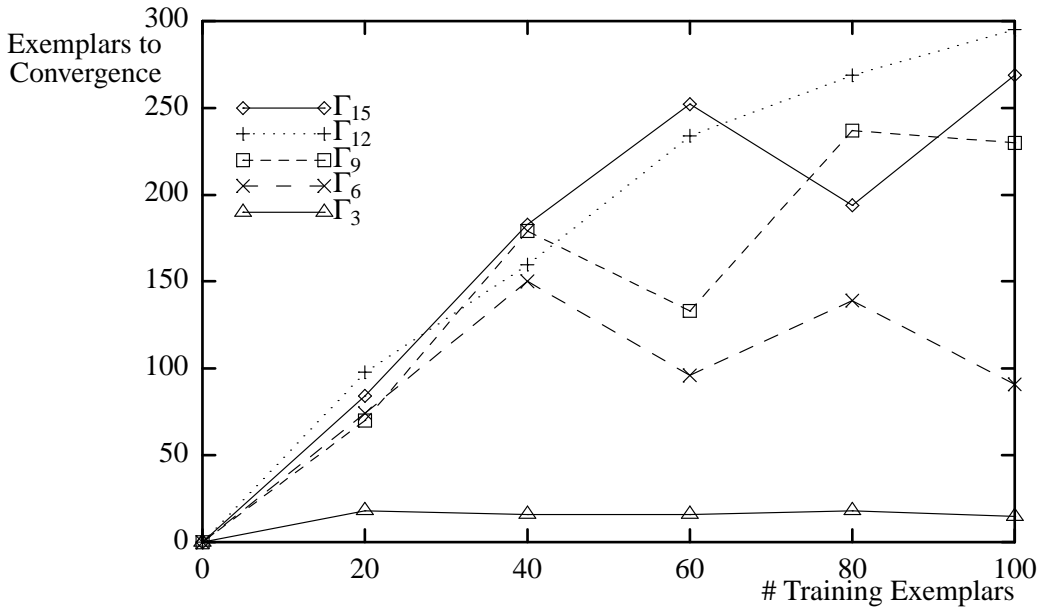

Figure 17: Number of exemplars processed until convergence for $\Gamma_i$ ($i = 3, 6, 9, 12, 15$).

Next we wish show the effects of positive bias, i.e., to show that user-provided guidance in the choice of initial weights can improve speed of convergence and accuracy in cross-validation. For each of the flawed theories $\Gamma_3$ and $\Gamma_{15}$, we compare the performance of PTR using default initial weights and biased initial weights ($\beta = 2$). In Figure 18, we show how cross-validation accuracy increases when bias is introduced. In Figure 19, we show how the number of examples which need to be processed until convergence decreases when bias is introduced.

Returning to the example above, we see that the introduction of bias allows PTR to immediately find the flawed clause $A \leftarrow p_6$ and to delete it straight away. In fact, PTR never requires the processing of more than 8 exemplars to do so. Thus, in this case, the introduction of bias both speeds up the revision process and results in the consistent choice of the optimal revision.

Moreover, it has also been shown in (Feldman, 1993) that PTR is robust with respect to random perturbations in the initial weights. In particular, in tests on thirty different synthetically-generated theories, introducing small random perturbations to each edge of a dt-graph before training resulted in less than 2% of test examples being classified differently than when training was performed using the original initial weights.

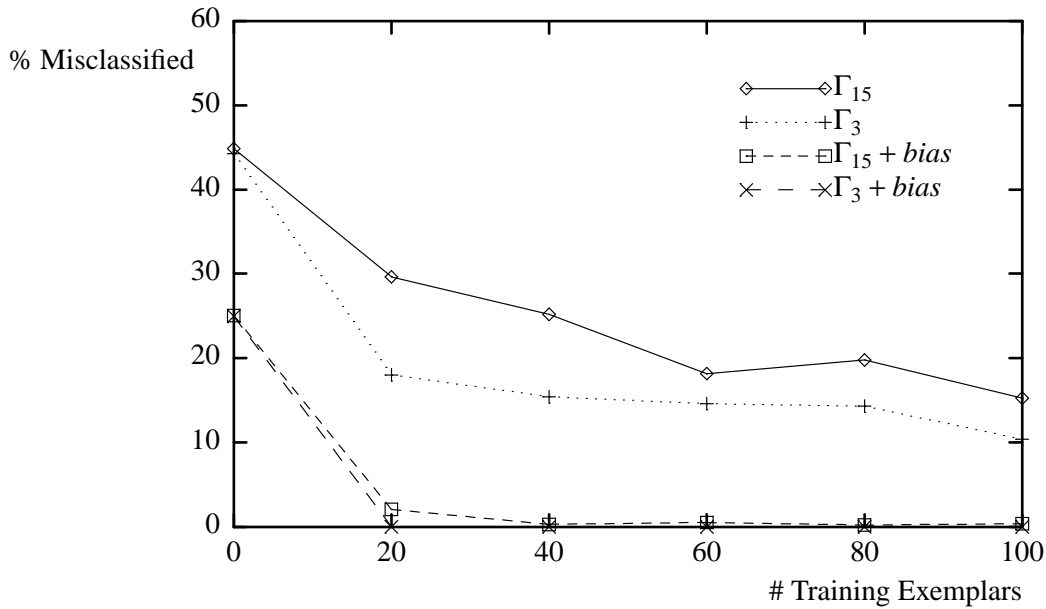

Figure 18: Error rates for the output theories produced by PTR from $\Gamma_i$ ($i = 3, 6, 9, 12, 15$), using favorably-biased initial weights.

## 6.4. Summary

Repairing internal literals and clauses is as natural for PTR as repairing leaves. Moreover, PTR converges rapidly. As a result, PTR scales up to deep theories without difficulty. Even for very badly flawed theories, PTR quickly finds repairs which correctly classify all known exemplars. These repairs are typically *less* radical than restoring the original theory and are close enough to the original theory to generalize accurately to test examples.

Moreover, although PTR is robust with respect to initial weights, user guidance in choosing these weights can significantly improve both speed of convergence and cross-validation accuracy.

## 7. Conclusions

In this paper, we have presented our approach, called PTR, to the theory revision problem for propositional theories. Our approach uses probabilities associated with domain theory elements to numerically track the "flow" of proof through the theory, allowing us to efficiently locate and repair flawed elements of the theory. We prove that PTR converges to a theory which correctly classifies all examples, and show experimentally that PTR is fast and accurate even for deep theories.

There are several ways in which PTR can be extended.

*First-order theories*. The updating method at the core of PTR assumes that provided exemplars unambiguously assign truth values to each observable proposition. In first-order theory revision the truth of an observable predicate typically depends on variable assignments.

194

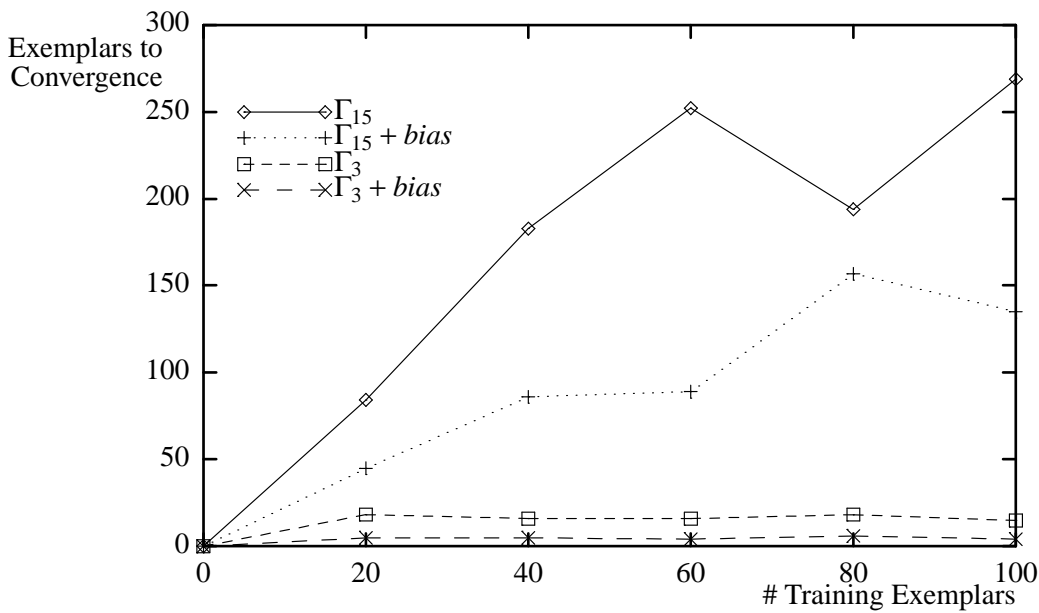

Figure 19: Number of exemplars processed until convergence using favorably-biased initial weights.

Thus, in order to apply PTR to first-order theory revision it is necessary to determine "optimal" variable assignments on the basis of which probabilities can be updated. One method for doing so is discussed in (Feldman, 1993).

*Inductive bias.* PTR uses bias to locate flawed elements of a theory. Another type of bias can be used to determine which revision to make. For example, it might be known that a particular clause might be missing a literal in its body but should under no circumstances be deleted, or that only certain types of literals can be added to the clause but not others. Likewise, it might be known that a particular literal is replaceable but not deletable, etc. It has been shown (Feldman *et al.*, 1993) that by modifying the inductive component of PTR to account for such bias, both convergence speed and cross-validation accuracy are substantially improved.

*Noisy exemplars.* We have assumed that it is only the domain theory which is in need of revision, but that the exemplars are all correctly classified. Often this is not the case. Thus, it is necessary to modify PTR to take into account the possibility of reclassifying exemplars on the basis of the theory rather than vice-versa. The PTR* algorithm (Section 6) suggests that misclassed exemplars can sometimes be detected before processing. Briefly, the idea is that an example which allows multiple proofs of some root is almost certainly IN for that root regardless of the classification we have been told. Thus, if $u_E(e_r)$ is high, then $E$ is probably IN regardless of what we are told; analogously, if $u_E(e_r)$ is low. A modified version of PTR based on this observation has already been successfully implemented (Koppel *et al.*, 1993).

In conclusion, we believe the PTR system marks an important contribution to the domain theory revision problem. More specifically, the primary innovations reported here are:

(1)  By assigning bias in the form of the probability that an element of a domain theory is flawed, we can clearly define the objective of a theory revision algorithm.

(2)  By reformulating a domain theory as a weighted dt-graph, we can numerically trace the flow of a proof or refutation through the various elements of a domain theory.

(3)  Proof flow can be used to efficiently update the probability that an element is flawed on the basis of an exemplar.

(4)  By updating probabilities on the basis of exemplars, we can efficiently locate flawed elements of a theory.

(5)  By using proof flow, we can determine precisely on the basis of which exemplars to revise a flawed element of the theory.

## Acknowledgments

The authors wish to thank Hillel Walters of Bar-Ilan University for his significant contributions to the content of this paper. The authors also wish to thank the JAIR reviewers for their exceptionally prompt and helpful remarks. Support for this research was provided in part by the Office of Naval Research through grant N00014-90-J-1542 (AMS, RF) and the Air Force Office of Scientific Research under contract F30602-93-C-0018 (AMS).

## Appendix A: Assigning Initial Weights

In this appendix we give one method for assigning initial weights to the elements of a domain theory. The method is based on the topology of the domain theory and assumes that no user-provided information regarding the likelihood of errors is available. If such information is available, then it can be used to override the values determined by this method.

The method works as follows. First, for each edge $e$ in $\Delta_\Gamma$ we define the "semantic impact" of $e$, $M(e)$. $M(e)$ is meant to signify the proportion of examples whose classification is directly affected by the presence of $e$ in $\Delta_\Gamma$.

One straightforward way of formally defining $M(e)$ is the following. Let $K^I$ be the pair $\langle \Delta_\Gamma, I \rangle$ such that $I$ assigns all root and negation edges the weight 1 and all other edges the weight $\frac{1}{2}$. Let $I(e)$ be identical to $I$ except that $e$ and all its ancestor edges have been assigned the weight 1. Let $E$ be the example such that for each observable proposition $P$ in $\Gamma$, $E(P)$ is the a priori probability that $P$ is true in a randomly selected example.[15] In particular, for the typical case in which observable propositions are Boolean and all example are equiprobable, $E(P) = \frac{1}{2}$. $E$ can be thought of as the "average" example. Then, if no edge of $\Delta_\Gamma$ has more than one parent-edge, we formally define the semantic significance, $M(e)$, of an edge $e$ in $\Delta_\Gamma$ as follows:

$$M(e) = u_E^{K^{I(e)}}(e_r) - u_E^{K_{\bar{e}}^{I(e)}}(e_r).$$

That is, $M(e)$ is the difference of the flow of $E$ through the root $r$, with and without the edge $e$.

Note that $M(e)$ can be efficiently computed by first computing $u_E^{K^I}(e)$ for every edge $e$ in a single bottom-up traversal of $\Delta_\Gamma$, and then computing $M(e)$ for every edge $e$ in a single top-down traversal of $\Delta_\Gamma$, as follows:

(1)  For a root edge $r$, $M(r) = 1 - u_E^{K^I}(r)$.

(2)  For all other edges, $M(e) = M(f(e)) \times \dfrac{2(1 - u_E^{K^I}(e))}{u_E^{K^I}(e)}$, where $f(e)$ is the parent edge of $e$.

If some edge in $\Delta_\Gamma$ has more than one parent-edge then we define $M(e)$ for an edge by using this method of computation, where in place of $M(f(e))$ we use $\max_f \left[ M(f(e)) \right]$.

Finally, for a set, $R$, of edges in $G$, we define $M(R) = \sum_{e \in R} M(e)$.[16]

Now, having computed $M(e)$ we compute the initial weight assignment to $e$, $p(e)$, in the following way. Choose some large $C$.[17] For each $e$ in $\Delta_\Gamma$ define:

---

[15] Although we have defined an example as a $\{0, 1\}$ truth assignment to each observable proposition, we have already noted in Footnote 4 that we can just as easily process examples which assign to observables any value in the interval $[0, 1]$.

[16] Observe that the number of examples reclassified as a result of edge-deletion is, in fact, superadditive, a fact not reflected by this last definition.

[17] We have not tested how to choose $C$ "optimally." In the experiments reported in Section 6, the value $C = 10^6$ was used.

$$p(e) = \frac{C^{\mathrm{M}(e)}}{C^{\mathrm{M}(e)} + 1}.$$

Now, regardless of how $\mathrm{M}(e)$ is defined, the virtue of this method of computing $p(e)$ from $\mathrm{M}(e)$ is the following: for such an initial assignment, $p$, *if two sets of edges $\langle \Delta_\Gamma, p \rangle$ are of equal total strength then as revision sets they are of equal radicality*. This means that all revision sets of equal strength are a priori equally probable.

For a set of edges of $\Delta_\Gamma$, define

$$S(e) = \begin{cases} 1 & \textit{if } e \in S \\ 0 & \textit{if } e \notin S \end{cases}$$

Then the above can be formalized as follows:

**Theorem A1**: If $R$ and $S$ are sets of elements of $\Gamma$ such that $\mathrm{M}(R) = \mathrm{M}(S)$ then it follows that $Rad(R) = Rad(S)$.

**Proof of Theorem A1**: Let $R$ and $S$ be sets of edges such that $\mathrm{M}(R) = \mathrm{M}(S)$. Recall that

$$Rad(S) = -\log\left( \prod_{e \,\in\, \Delta} [1 - p(e)]^{S(e)} \times [p(e)]^{1 - S(e)} \right)$$

Then

$$\frac{\exp(-Rad(S))}{\exp(-Rad(R))} = \prod_{e \,\in\, \Delta} \frac{[1 - p(e)]^{S(e)} \times p(e)^{1 - S(e)}}{[1 - p(e)]^{R(e)} \times p(e)^{1 - R(e)}}$$

$$= \prod_{e \,\in\, \Delta} \left[ p(e) 1 - p(e) \right]^{R(e) - S(e)}$$

$$= \prod_{e \,\in\, \Delta} \left[ C^{\mathrm{M}(e)} \right]^{R(e) - S(e)}$$

$$= C^{\mathrm{M}(R) - \mathrm{M}(S)} = 1.$$

It follows immediately that $Rad(R) = Rad(S)$. $\square$

A simple consequence which illustrates the intuitiveness of this theorem is the following: suppose we have two possible revisions of $\Delta$, each of which entails deleting a simple literal. Suppose further that one literal, $l_1$, is deep in the tree and the other, $l_2$, is higher in the tree so that $\mathrm{M}(l_2) = 4 \times \mathrm{M}(l_1)$. Then, using default initial weights as assigned above, the radicality of deleting $l_2$ is 4 times as great as the radicality of deleting $l_1$.

## Appendix B: Updated Weights as Conditional Probabilities

In this appendix we prove that under certain limiting conditions, the algorithm computes the conditional probabilities of the edges given the classification of the example.

Our first assumption for the purpose of this appendix is that the correct dt-graph $\Delta_\Theta$ is known to be a subgraph of the given dt-graph $\Delta_\Gamma$. This means that for every node $n$ in $\Delta_\Gamma$, $w(n) = 1$ (and, consequently, for every edge $e$ in $\Delta_\Gamma$, $p(e) = w(e)$). A pair $\langle \Delta_\Gamma, w \rangle$ with this property is said to be *deletion-only*.

Although we informally defined probabilities directly on edges, for the purposes of this appendix we formally define our probability function on the space of all subgraphs of $\Delta_\Gamma$. That is, the elementary events are of the form $\Delta_\Theta = \Delta_{\Gamma'}$ where $\Delta_{\Gamma'} \subseteq \Delta_\Gamma$. Then the probability that $e \in \Delta_\Theta$ is simply $\sum_{\Gamma' \subseteq \Gamma} \{ p(\Delta_\Theta = \Delta_{\Gamma'}) | e \in \Delta_{\Gamma'} \}$.

We say that a deletion-only, weighted dt-graph $\langle \Delta_\Gamma, p \rangle$ is *edge-independent* if for any $\Gamma' \subseteq \Gamma$,

$$p(\Delta_\Theta = \Delta_{\Gamma'}) = \prod_{e \in \Delta_{\Gamma'}} p(e) \times \prod_{e \notin \Delta_{\Gamma'}} 1 - p(e).$$

Finally, we say that $\Delta_\Gamma$ is *tree-like* if no edge $e \in \Delta_\Gamma$ has more than one parent-edge. Observe that any dt-graph which is connected and tree-like has only one root.

We will prove results for deletion-only, edge-independent, tree-like weighted dt-graphs.[18]

First we introduce some more terminology. Recall that every node in $\Delta_\Gamma$ is labeled by one of the literals in $\hat{\Gamma}$ and that by definition, this literal is true if not all of its children in $\Delta_{\hat{\Gamma}}$ are true. Recall also that the dt-graph $\Delta_{\Gamma'} \subseteq \Delta_\Gamma$ represents the sets of NAND equations, $\hat{\Gamma}' \subseteq \hat{\Gamma}$. A literal $l$ in $\hat{\Gamma}$ forces its parent in $\hat{\Gamma}$ to be true, given the set of equations $\hat{\Gamma}'$ and the example $E$, if $l$ appears in $\hat{\Gamma}'$ and is false given $\hat{\Gamma}'$ and $E$. (This follows from the definition of NAND.) Thus we say that an edge $e$ in $\Delta_\Gamma$ is *used* by $E$ in $\Delta_{\Gamma'}$ if $e \in \Delta_{\Gamma'}$ and $\hat{\Gamma}' \vdash_E \neg n_e$.

If $e$ is not used by $E$ in $\Delta_{\Gamma'}$ we write $N_E^{\Gamma'}(e)$. Note that $N_E^{\Gamma'}(e_r)$ if and only if $\Gamma'(E) = 1$.

Note that, given the probabilities of the elementary events $\Delta_{\Gamma'} = \Delta_\Theta$, the probability $p(N_E^\Theta(e))$ that the edge $e$ is not used by $E$ in the target domain theory $\Theta$ is simply $\sum_{\Gamma' \subseteq \Gamma} \left\{ p(\Delta_{\Gamma'} = \Delta_\Theta) | N_E^{\Gamma'}(e) \right\}$. Where there is no ambiguity we will use $N_E(e)$ to refer to $N_E^\Theta(e)$.

**Theorem B1**: If $\langle \Delta_\Gamma, w \rangle$ is a deletion-only, edge-independent, tree-like weighted dt-graph, then for every edge $e$ in $\Delta_\Gamma$, $u_E(e) = p(N_E(e))$.

**Proof of Theorem B1**: We use induction on the distance of $n_e$ from its deepest descendant. If $n_e$ is an observable proposition $P$ then $e$ is used by $E$ in $\Theta$ precisely if $e \in \Theta$ and $P$ is false in $E$. Thus the probability that $e$ is not used by $E$ in $\Theta$ is $[1 - p(e)] \times [1 - E(P)] = u_E(e)$.

---

[18] Empirical results show that our algorithm yields reasonable approximations of the conditional probabilities even when these conditions do not hold.

If $n_e$ is not a observable proposition then $\hat{\Theta} \models_E \neg n_e$ precisely if all its children $\hat{\Theta}$ are true in $\hat{\Theta}$, that is, if all its children are unused in $\hat{\Theta}$. But then

$$p(N_E(e)) = p(e) \times p(\Theta \models_E \neg n_e) \qquad \text{(edge independence)}$$

$$= p(e) \times \prod_{s \in children(e)} p(N_E(s)) \qquad \text{(induction hypothesis)}$$

$$= p(e) \times \prod_{s \in children(e)} u_E(s)$$

$$= u_E(e).$$

□

This justifies the bottom-up part of the algorithm. In order to justify the top-down part we need one more definition.

Let $p(e | \langle E, \Theta(E) \rangle)$ be the probability that $e \in \Delta_\Theta$ given $\langle \Delta_\Gamma, p \rangle$ and the exemplar $\langle E, \Theta(E) \rangle$. Then

$$p(e | \langle E, \Theta(E) \rangle) = \frac{\displaystyle\sum_{\Gamma' \subseteq \Gamma} \{p(\Delta_\Theta = \Delta_{\Gamma'}) | e \in \Delta_{\Gamma'}, \Theta(E) = \Gamma'(E)\}}{\displaystyle\sum_{\Gamma' \subseteq \Gamma} \{p(\Delta_\Theta = \Delta_{\Gamma'}) | \Theta(E) = \Gamma'(E)\}} .$$

Now we have

**Theorem B2**: If $\langle \Delta_\Gamma, w \rangle$ is deletion-only, edge-independent and tree-like, then for every edge $e$ in $\Delta_\Gamma$, $p_{new}(e) = p(e | \langle E, \Theta(E) \rangle)$.

In order to prove the theorem we need several lemmas:

**Lemma B1**: For every example $E$ and every edge $e$ in $\Delta_\Gamma$

$$p(\neg N_E(e)) = p(\neg N_E(e), N_E(f(e))) = p(\neg N_E(e) | N_E(f(e))) \times p(N_E(f(e))).$$

This follows immediately from the fact that if an edge, $e$, is used, then its parent-edge, $f(e)$, is not used.

**Lemma B2**: For every example $E$ and every edge $e$ in $\Delta_\Gamma$,

$$p(N_E(E) | N_E(f(e)), \langle E, \Theta(E) \rangle) = p(N_E(e) | N_E(f(e))).$$

This lemma states that $N_E(e)$ and $\langle E, \Theta(E) \rangle$ are conditionally independent given $N_E(f(e))$ (Pearl, 1988). That is, once $N_E(f(e))$ is known, $\langle E, \Theta(E) \rangle$ adds no information regarding $N_E(e)$. This is immediate from the fact that $p(\langle E, \Theta(E) \rangle | N_E(f(e)))$ can be expressed in terms of the probabilities associated with non-descendants of $f(e)$, while $p(N_E(e))$ can be expressed in terms of the probabilities associated with descendants of $r(e)$.

**Lemma B3**: For every example $E$ and every edge $e$ in $\Delta_\Gamma$,

$$v_E(e) = p(N_E(e) | \langle E, \Theta(E) \rangle).$$

**Proof of Lemma B3**: The proof is by induction on the depth of the edge, $e$. For the root edge, $e_r$, we have

$$v_E(e_r) = \Theta(E) = p(\Theta(E) = 1 | \langle E, \Theta(E) \rangle) = p(N_E(e_r) | \langle E, \Theta(E) \rangle).$$

Assuming that the theorem is known for $f(e)$, we show that it holds for $e$ as follows:

$$1 - v_E(e) = \left[ 1 - u_E(e) \right] \frac{v_E(f(e))}{u_E(f(e))} \qquad \text{(definition of } v)$$

$$= p(\neg N_E(e)) \times \frac{v_E(f(e))}{p(N_E(f(e)))} \qquad \text{(Theorem B1)}$$

$$= p(N_E(e) | \langle E, \Theta(E) \rangle) \times \frac{p(\neg N_E(e))}{p(N_E(f(e)))} \qquad \text{(induction hypothesis)}$$

$$= p(N_E(e) | \langle E, \Theta(E) \rangle) \times p(\neg N_E(e) | N_E(f(e))) \qquad \text{(Lemma B1)}$$

$$= p(N_E(e) | \langle E, \Theta(E) \rangle) \\ \times p(\neg N_E(e) | N_E(f(e)), \langle E, \Theta(E) \rangle) \qquad \text{(Lemma B2)}$$

$$= p(\neg N_E(e), N_E(f(e)) | \langle E, \Theta(E) \rangle) \qquad \text{(Bayes rule)}$$

$$= p(\neg N_E(e) | \langle E, \Theta(E) \rangle) \qquad \text{(Lemma B1)}$$

$$= 1 - p(N_E(e) | \langle E, \Theta(E) \rangle).$$

$\square$

Let $\neg e$ be short for the event $e \notin \Delta_\Theta$. Then we have

**Lemma B4**: For every example $E$ and every edge $e$ in $\Delta_\Gamma$,

$$p(\neg e) = p(\neg e, \neg N_E(e)) = p(\neg e | N_E(e)) \times p(N_E(e)).$$

This lemma, which is analogous to Lemma B1, follows from the fact that if $e$ is deleted, then $e$ is unused.

**Lemma B5**: For every example $E$ and every edge $e$ in $\Delta_\Gamma$,

$$p(\neg e | \neg N_E(e), \langle E, \Theta(E) \rangle) = p(\neg e | \neg N_E(e)).$$

This lemma, which is analogous to Lemma B2, states that $\neg e$ and $\langle E, \Theta(E) \rangle$ are conditionally independent given $\neg N_E(e)$. That is, once $\neg N_E(e)$ is known, $\langle E, \Theta(E) \rangle$ adds no information regarding the probability of $\neg e$. This is immediate from the fact that $p(\langle E, \Theta(E) \rangle | \neg N_E(e))$ can be expressed in terms of the probabilities of edges other than $e$.

We now have all the pieces to prove Theorem B2.

**Proof of Theorem B2**:

$$1 - p_{new}(e) = \left[ 1 - p(e) \right] \frac{v_E(e)}{u_E(e)} \qquad \text{(definition of } p_{new})$$

$$= p(\neg e) \times \frac{v_E(e)}{p(N_E(e))} \qquad \text{(Theorem B1)}$$

$$= p(N_E(e)|\langle E, \Theta(E)\rangle) \times \frac{p(\neg e)}{p(N_E(e))} \qquad \text{(Lemma B3)}$$

$$= p(N_E(e)|\langle E, \Theta(E)\rangle) \times p(\neg e|N_E(e)) \qquad \text{(Lemma B4)}$$

$$= p(N_E(e)|\langle E, \Theta(E)\rangle) \times p(\neg e|N_E(e), \langle E, \Theta(E)\rangle) \qquad \text{(Lemma B5)}$$

$$= p(\neg e, N_E(e)|\langle E, \Theta(E)\rangle) \qquad \text{(Bayes rule)}$$

$$= p(\neg e|\langle E, \Theta(E)\rangle) \qquad \text{(Lemma B4)}$$

$$= 1 - p(e|\langle E, \Theta(E)\rangle).$$

$\square$

## Appendix C: Proof of Convergence

We have seen in Section 5 that PTR always terminates. We wish to show that when it does, all exemplars are classified correctly. We will prove this for domain theories which satisfy certain conditions which will be made precise below. The general idea of the proof is the following: by definition, the algorithm terminates either when all exemplars are correctly classified or when all edges have weight 1. Thus, it is only necessary to show that it is not possible to reach a state in which all edges have weight 1 and some exemplar is misclassified. We will prove that such a state fails to possess the property of "consistency" which is assumed to hold for the initial weighted dt-graph K, and which is preserved at all times by the algorithm.

**Definition** (*Consistency*): The weighted dt-graph $K = \langle \Delta, p \rangle$ is *consistent* with exemplar $\langle E, \Theta(E) \rangle$ if, for every root $r_i$ in $\Delta$, either:

 (i)  $\Theta_i(E) = 1$ and $u_E^K(r_i) > 0$, or
 (ii)  $\Theta_i(E) = 0$ and $u_E^K(r_i) < 1$.

Recall that an edge $e$ is defined to be even if it is of even depth along every path from a root and odd if is of odd depth along every path from a root. A domain theory is said to be *unambiguous* if every edge is either odd or even. Note that negation-free domain theories are unambiguous. We will prove our main theorem for unambiguous, single-root domain theories.

 Recall that the only operations performed by PTR are:

 (1)  updating weights,
 (2)  deleting even edges,
 (3)  deleting odd edges,
 (4)  adding a subtree beneath an even edge, and
 (5)  adding a subtree beneath an odd edge.

We shall show that each of these operations is performed in such a way as to preserve consistency.

**Theorem C1** (*Consistency*): If $K = \langle \Delta, p \rangle$ is a single-rooted, unambiguous weighted dt-graph which is consistent with the exemplar $\langle E, \Theta(E) \rangle$ and $K' = \langle \Delta', p' \rangle$ is obtained from K via a single operation performed by PTR, then K' is also a single-rooted, unambiguous dt-graph which is consistent with $E$.

Before we prove this theorem we show that it easily implies convergence of the algorithm.

**Theorem C2** (*Convergence*): Given a single-rooted, unambiguous weighted dt-graph K and a set of exemplars Z such that K is consistent with every exemplar in Z, PTR terminates and produces a dt-graph $\Delta'$ which classifies every exemplar in Z correctly.

**Proof of Theorem C2**: If PTR terminates prior to each edge being assigned the weight 1, then by definition, all exemplars are correctly classified. Suppose then that PTR produces a weighted dt-graph $K' = \langle \Delta', p' \rangle$ such that $p'(e) = 1$ for every $e \in \Delta'$. Assume, contrary to the theorem, that some exemplar $\langle E, \Theta(E) \rangle$ is misclassified by K' for the root $r$. Without loss of generality, assume that $\langle E, \Theta(E) \rangle$ is an IN exemplar of $r$. Since $p'(e) = 1$ for every edge, this means that $u_E^{K'}(e_r) = 0$. But this is impossible since the consistency of K implies that $u_E^K(e_r) > 0$ and thus it follows from Theorem C1 that for any K' obtainable form

K, $u_E^{K'}(e_r) > 0$. This contradicts the assumption that $E$ is misclassified by K'. □

Let us now turn to the proof of Theorem C1. We will use the following four lemmas, slight variants of which are proved in (Feldman, 1993).

**Lemma C1**: If K' $= \langle \Delta, p' \rangle$ is obtained from K $= \langle \Delta, p \rangle$ via updating of weights, then for every edge $e \in \Delta$ such that $0 < p(e) < 1$, we have $0 < p'(e) < 1$.[19]

**Lemma C2**: Let K $= \langle \Delta, p \rangle$ be a weighted dt-graph such that $0 < u_E^K(e_r) < 1$ and let K' $= \langle \Delta, p' \rangle$. Then if for every edge $e$ in $\Delta$ such that $0 < p(e) < 1$, we have $0 < p'(e) < 1$, it follows that $0 < u_E^K(e_r) < 1$.

**Lemma C3**: Let K $= \langle \Delta, p \rangle$ be a weighted dt-graph such that $u_E^K(e_r) > 0$ and let K' $= \langle \Delta', p' \rangle$. The, if for every edge $e$ in $\Delta$, it holds that either:

   (i)   $p'(e) = p(e)$, or
   (ii)  $depth(e)$ is odd and $u_E^{K'}(e) > 0$, or
   (iii) $depth(e)$ is even and $u_E^{K'}(e) < 1$

then $u_E^{K'}(e) > 0$.

An analogous lemma holds where the roles of "$> 0$" and "$< 1$" are reversed.

**Lemma C4**: If $e$ is even edge in K, then $u_E^{K_{\bar{e}}}(e_r) \geq u_E^K(e_r) \geq u_E^{K_e}(r)$. In addition, if $e$ is an odd edge in K, then $u_E^{K_{\bar{e}}}(e_r) \leq u_E^K(e_r) \leq u_E^{K_e}(r)$.

We can now prove consistency (Theorem C1). We assume, without loss of generality, that $\langle E, \Theta(E) \rangle$ is an IN exemplar of the root $r$ and prove that for each one of the five operations (updating and four revision operators) of PTR, that if K' is obtained by that operation from K and $u_E^K(e_r) > 0$, then $u_E^{K'}(e_r) > 0$.

**Proof of Theorem C1**: The proof consists of five separate cases, each corresponding to one of the operations performed by PTR.

**Case 1**: K' is obtained from K via updating of weights.

By Lemma C1, for every edge $e$ in $\Delta$, if $0 < p(e) < 1$ then $0 < p'(e) < 1$. But then by Lemma C2, if $u_E^K(e_r) > 0$ then $u_E^{K'}(e_r) > 0$.

**Case 2**: K' is obtained from K via deletion of an even edge, $e$.

From Lemma C4(i), we have $u_E^{K_{\bar{e}}}(e_r) \geq u_E^K(e_r) > 0$.

**Case 3**: K' is obtained from K via deletion of an odd edge, $e$.

The edge $e$ is deleted only if it is not needed for any exemplar. Suppose that, contrary to the theorem, there is an IN exemplar $\langle E, \Theta(E) \rangle$ such that $u_E^K(e_r) > 0$ but $u_E^{K'}(e_r) = 0$. Then

---

[19] Recall that in the updating algorithm we defined

$$v_E(e_{r_i}) = \begin{cases} \varepsilon & \text{if } \Theta_i(E) = 0 \\ 1 - \varepsilon & \text{if } \Theta_i(E) = 1 \end{cases}.$$

The somewhat annoying presence of $\varepsilon > 0$ is necessary for the proof of Lemma C1.

$$R(\langle E, \Theta(E) \rangle, e, \mathrm{K}) = \frac{u_E^{\mathrm{K}_e}(e_r)}{u_E^{\mathrm{K}_{\bar{e}}}(e_r)}$$

$$= \frac{u_E^{\mathrm{K}_e}(e_r)}{u_E^{\mathrm{K}'}(e_r)}$$

$$= \frac{u_E^{\mathrm{K}_e}(e_r)}{0} > 2.$$

But then $e$ is needed for $E$, contradicting the fact that $e$ is not needed for any exemplar.

**Case 4**: $\mathrm{K}'$ is obtained from $\mathrm{K}$ via appending a subtree beneath an even edge, $e$.

If $p'(e) < 1$, then the result is immediate from Lemma C2. Otherwise, let $f$ be the root edge of the subtree $\Delta_a$ which is appended to $\Delta$, beneath $e$. Then $\mathrm{K}'|f = \mathrm{K}_e$. Suppose that, contrary to the theorem, there is some IN exemplar $\langle E, \Theta(E) \rangle$ such that $u_E^{\mathrm{K}}(e_r) > 0$ but $u_E^{\mathrm{K}'}(e_r) = 0$. Then by Lemma C4(ii), $u_E^{\mathrm{K}_e}(e_r) = u_E^{\mathrm{K}'|e}(e_r) \leq u_E^{\mathrm{K}'}(e_r) = 0$. But then,

$$R(\langle E, \Theta(E) \rangle, e, \mathrm{K}) = \frac{u_E^{\mathrm{K}_e}(e_r)}{u_E^{\mathrm{K}_{\bar{e}}}(e_r)}$$

$$\leq \frac{0}{u_E^{\mathrm{K}_{\bar{e}}}(e_r)} = 0.$$

Thus $e$ is destructive for $E$ in $\mathrm{K}$. But then, by the construction of $\Delta_a$, $u_E^{\mathrm{K}'}(f) = 1$. Thus, $u_E^{\mathrm{K}'}(e) = 0 < 1$. The result follows immediately from Lemma C3.

**Case 5**: $\mathrm{K}'$ is obtained from $\mathrm{K}$ via appending a subtree to $\mathrm{K}$ beneath the odd edge, $e$.

Suppose that, contrary to the theorem, some IN exemplar $\langle E, \Theta(E) \rangle$, $u_E^{\mathrm{K}}(e_r) > 0$ but $u_E^{\mathrm{K}'}(e_r) = 0$. Since $\mathrm{K}'_{\bar{e}} = \mathrm{K}_{\bar{e}}$, it follows that

$$R(\langle E, \Theta(E) \rangle, e, \mathrm{K}) = \frac{u_E^{\mathrm{K}_e}(e_r)}{u_E^{\mathrm{K}_{\bar{e}}}(e_r)}$$

$$= \frac{u_E^{\mathrm{K}_e}(e_r)}{u_E^{\mathrm{K}'_{\bar{e}}}(e_r)}.$$

Now, using Lemma C4(ii) on both numerator and denominator, we have

$$\frac{u_E^{\mathrm{K}_e}(e_r)}{u_E^{\mathrm{K}'_{\bar{e}}}(e_r)} \geq u_E^{\mathrm{K}}(e_r) u_E^{\mathrm{K}'}(e_r) = \infty > 2.$$

Thus, $e$ is needed for $E$ in $\mathrm{K}$. Now, let $f$ be the root edge of the appended subtree, $\Delta_a$. Then, by the construction of $\Delta_a$, it follows that $u_E^{\mathrm{K}'}(f) < 1$ and, therefore $u_E^{\mathrm{K}'}(e) > 0$. The result is immediate from Lemma C3.

This completes the proof of the theorem. □

It is instructive to note why the proof of Theorem C1 fails if $\Delta$ is not restricted to unambiguous single-rooted dt-graphs. In case 4 of the proof of Theorem C1, we use the fact that if an edge $e$ is destructive for an exemplar $\langle E, \Theta(E) \rangle$ then the revision algorithm used to construct the subgraph, $\Delta_a$, appended to $e$ will be such that $u_E^{K'}(f) = 1$. However, this fact does not hold in the case where $e$ is simultaneously needed and destructive. This can occur if $e$ is a descendant of two roots where $E$ is IN for one root and OUT for another root. It can also occur when one path from $e$ to the root $r$ is of even length and another path is of odd length.

## Appendix D: Guide to Notation

| | |
|---|---|
| $\Gamma$ | A domain theory consisting of a set of clauses of the form $C_i\colon H_i \leftarrow B_i$. |
| $C_i$ | A clause label. |
| $H_i$ | A clause head; it consists of a single positive literal. |
| $B_i$ | A clause body; it consists of a conjunction of positive or negative literals. |
| $E$ | An example; it is a set of observable propositions. |
| $\Gamma_i(E)$ | The classification of the example $E$ for the $i$th root according to domain theory $\Gamma$. |
| $\Theta_i(E)$ | The correct classification of the example $E$ for the $i$th root. |
| $\langle E, \Theta(E) \rangle$ | An exemplar, a classified example. |
| $\hat{\Gamma}$ | The set of NAND clauses equivalent to $\Gamma$. |
| $\Delta_\Gamma$ | The dt-graph representation of $\Gamma$. |
| $n_e$ | The node to which the edge $e$ leads. |
| $n^e$ | The node from which the edge $e$ comes. |
| $p(e)$ | The weight of the edge $e$; it represents the probability that the edge $e$ needs to be deleted or that edges need to be appended to the node $n_e$. |
| $\mathrm{K} = \langle \Delta, p \rangle$ | A weighted dt-graph. |
| $\mathrm{K}_e$ | Same as K but with the weight of the edge $e$ equal to 1. |
| $\mathrm{K}_{\bar{e}}$ | Same as K but with the edge $e$ deleted. |
| $u_E(e)$ | The "flow" of proof from the example $E$ through the edge $e$. |
| $v_E(e)$ | The adjusted flow of proof through $e$ taking into account the correct classification of the example $E$. |
| $R_i(\langle E, \Theta(E) \rangle, e, \mathrm{K})$ | The extent (ranging from 0 to $\infty$) to which the edge $e$ in the weighted dt-graph K contributes to the correct classification of the example $E$ for the $i$th root. If $R_i$ is less/more than 1, then $e$ is harmful/helpful; if $R_i = 1$ then $e$ is irrelevant. |
| $\sigma$ | The revision threshold; if $p(e) < \sigma$ then $e$ is revised. |
| $\lambda$ | The weight assigned to a revised edge and to the root of an appended component. |
| $\delta_\sigma$ | The revision threshold increment. |
| $\delta_\lambda$ | The revised edge weight increment. |
| $Rad_{\mathrm{K}}(\Gamma')$ | The radicality of the changes required to K in order to obtain a revised theory $\Gamma'$. |